\documentclass[runningheads]{llncs}

\usepackage{eccv}

\usepackage{eccvabbrv}

\usepackage{graphicx}
\usepackage{booktabs}

\usepackage[accsupp]{axessibility}  %

\usepackage[pagebackref,breaklinks,colorlinks,citecolor=eccvblue]{hyperref}

\usepackage{orcidlink}
\usepackage{booktabs}

\usepackage[dvipsnames]{xcolor}

\usepackage{graphicx}
\usepackage{subcaption}
\usepackage{float}
\usepackage{caption}
\usepackage{lscape}                                         %
\usepackage{wrapfig}

\usepackage[lined,ruled,linesnumbered]{algorithm2e}
\usepackage{animate} %

\usepackage{booktabs}                   %
\usepackage{multirow}

\usepackage{enumitem}

\usepackage{bm}                          %

\usepackage{amssymb}
\usepackage{amsmath}  %

\usepackage{units}
\usepackage{color}
\usepackage[T1]{fontenc}    %
\usepackage{amsfonts}       %
\usepackage[utf8]{inputenc} %
\usepackage{nicefrac}       %
\usepackage{microtype}      %
\usepackage{pifont}         %
\usepackage{comment}

\usepackage[mathscr]{euscript}
\usepackage{colortbl}

\usepackage{url}  %

\usepackage{listings}
\usepackage{xcolor}
\usepackage{cuted}

\def\ie{\textit{i.e.,\ }}
\def\eg{\textit{e.g.,\ }}
\def\etal{\textit{et~al.\ }}

\newcommand{\mycaption}[2]{\caption{\textbf{#1}. #2}}

\newcommand{\ignore}[1]{}   %

\definecolor{codegreen}{rgb}{0,0.6,0}
\definecolor{codegray}{rgb}{0.5,0.5,0.5}
\definecolor{codepurple}{rgb}{0.58,0,0.82}
\definecolor{backcolour}{rgb}{0.95,0.95,0.92}

\lstdefinestyle{style-python}{
    basicstyle=\ttfamily\scriptsize,
    backgroundcolor=\color{backcolour},   
    commentstyle=\color{codegreen},
    keywordstyle=\color{magenta},
    numberstyle=\tiny\color{codegray},
    stringstyle=\color{codepurple},
    breakatwhitespace=false,
    breaklines=true,
    captionpos=b,
    keepspaces=true,
    numbers=left,
    numbersep=5pt,
    showspaces=false,
    showstringspaces=false,
    showtabs=false,
    tabsize=2
}

\lstdefinestyle{style-prompt}{
    basicstyle=\ttfamily\scriptsize,
    backgroundcolor=\color{backcolour},
    basewidth=0.55em,
    commentstyle=\color{codegreen},
    breakatwhitespace=false,
    breaklines=true,
    captionpos=b,
    keepspaces=true,
    showspaces=false,
    showstringspaces=false,
    keywordstyle=\ttfamily,
    numberstyle=\tiny\color{codegray},
    numbers=left,
    numbersep=5pt, 
    moredelim=[is][\color{red}]{|}{|},
    moredelim=[is][\color{nice-blue}\bfseries]{//}{//},
    moredelim=[is][\color{nice-orange}\bfseries]{[[}{]]}
}

\colorlet{dark-blue}{blue!50!black}
\colorlet{dark-cyan}{cyan!75!black}
\colorlet{dark-purple}{purple!50!black}
\colorlet{dark-red}{red!75!black}
\colorlet{dark-green}{green!75!black}
\colorlet{dark-orange}{orange!50!black}
\colorlet{dark-gray}{black!75}
\colorlet{light-gray}{black!30}
\definecolor{nice-red}{HTML}{E41A1C}
\definecolor{nice-orange}{HTML}{FF7F00}
\definecolor{nice-yellow}{HTML}{FFC020}
\definecolor{nice-green}{HTML}{39b54a}
\definecolor{nice-blue}{HTML}{0071bc}
\definecolor{nice-purple}{HTML}{984EA3}
\definecolor{cvprblue}{rgb}{0.21,0.49,0.74}

\definecolor{darkGreen}{rgb}{0, 0.6, 0}
\definecolor{darkRed}{rgb}{0.9, 0, 0} 
\definecolor{cyan}{rgb}{0, 0.5, 0.6} 
\definecolor{darkViolet}{rgb}{0.58, 0, 0.83}
\definecolor{lightgreen}{rgb}{0.4, .9, 0.4}
\definecolor{lightred}{rgb}{1, 0.5, 0.51}
\definecolor{xgray}{rgb}{0.6, 0.6, 0.6}
\definecolor{Highlight}{HTML}{39b54a}
\definecolor{citecolor}{HTML}{0071bc}

\definecolor{bg-gray}{gray}{0.95}
\definecolor{neon-blue}{HTML}{00A2E8}
\definecolor{neon-green}{HTML}{b5E613}

\begin{document}

\title{Think-Program-reCtify: 3D Situated Reasoning with Large Language Models} 

\titlerunning{Think-Program-reCtify: 3D Situated Reasoning with Large Language Models}

\author{Qingrong He\inst{1} \and
    Kejun Lin\inst{1} \and
    Shizhe Chen\inst{2}\textsuperscript{*} \and 
    Anwen Hu\inst{3} \and 
    Qin Jin\inst{1}\textsuperscript{*}\textsuperscript{\dag}
}

\authorrunning{Q.~He et al.}

\institute{\textsuperscript{1}Renmin University of China, \textsuperscript{2}INRIA, \textsuperscript{3}Alibaba Group}

\maketitle

\footnotetext{\textsuperscript{*}Equal supervision}
\footnotetext{\textsuperscript{\dag}Corresponding author}

\begin{abstract}
This work addresses the 3D situated reasoning task which aims to answer questions given egocentric observations in a 3D environment.
The task remains challenging as it requires comprehensive 3D perception and complex reasoning skills. 
End-to-end models trained on supervised data for 3D situated reasoning suffer from data scarcity and generalization ability.
Inspired by the recent success of leveraging large language models (LLMs) for visual reasoning, we propose LLM-TPC, a novel framework that leverages the planning, tool usage, and reflection capabilities of LLMs through a \textbf{T}hink-\textbf{P}rogram-re\textbf{C}tify loop.  
The \textbf{Think} phase first decomposes the compositional question into a sequence of steps, and then the \textbf{Program} phase grounds each step to a piece of code and calls carefully designed 3D visual perception modules.
Finally, the \textbf{Rectify} phase adjusts the plan and code if the program fails to execute. 
Experiments and analysis on the SQA3D benchmark demonstrate the effectiveness, interpretability and robustness of our method.
Our code is publicly available at \url{https://qingrongh.github.io/LLM-TPC/}.
\end{abstract}

\section{Introduction}
\label{sec:intro}

\begin{figure}[t]
  \centering
  \vspace{+20pt}
    \includegraphics[width=\linewidth]{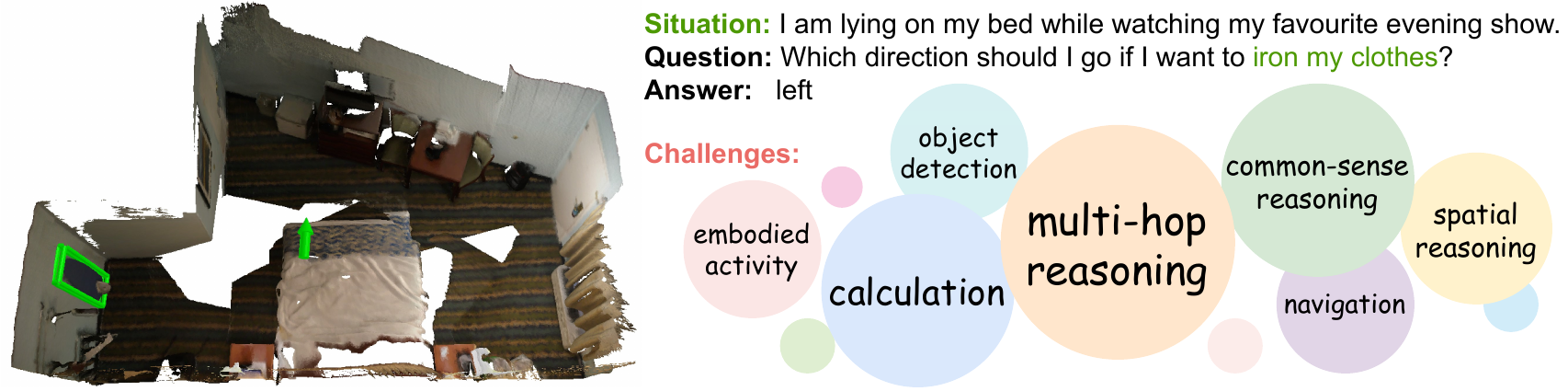}

    \vspace{+5pt}
   \caption{Situated reasoning in 3D scenes. It aims to answer complex questions given  egocentric situation in a 3D environment. The \textcolor{dark-green}{green arrow} indicates the position and orientation described by the situation, and the \textcolor{dark-green}{green box} refers to the target object.}
   \label{fig:3dsr}
\vspace{+10pt}
\end{figure}

Understanding and reasoning over 3D environments is a key capability for autonomous agents to assist humans in the real world~\cite{indoorsu}.
In this work, we focus on the 3D Situated Reasoning (3D-SR) task~\cite{sqa3d} as illustrated in \cref{fig:3dsr}, which aims to answer complex questions given the egocentric situation in a 3D environment.
It is a very challenging task as it requires comprehensive 3D vision perception and multiple reasoning skills.
For example, as shown in \cref{fig:3dsr}, to answer the question `Which direction should I go if I want to iron my clothes?', an agent should at least possess the following capabilities: (i) common-sense of object affordance for inferring the ironing board as the target object; (ii) open-vocabulary 3D object detection for localizing the ironing board in the scene; and (iii) 3D spatial reasoning for calculating the relative direction of the ironing board to its current location.

\begin{figure*}
    \centering
    \begin{subfigure}{0.28\linewidth}
    \includegraphics[width=\linewidth]{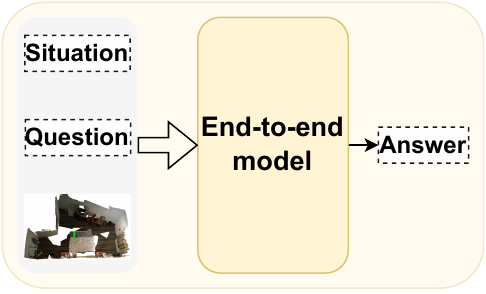}
    \caption{End-to-end methods.}
    \label{fig:end_to_end}
    \end{subfigure}
    \hfill
    \begin{subfigure}{0.28\linewidth}
    \includegraphics[width=\linewidth]{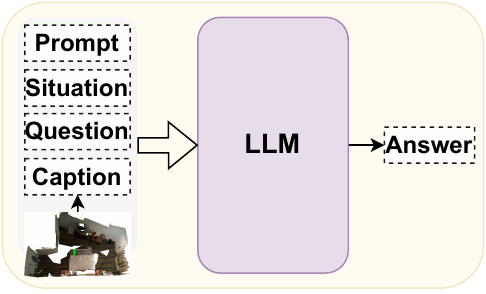}
    \caption{Language-only methods.}
    \label{fig:llm_caption}
    \end{subfigure}
    \hfill
    \begin{subfigure}{0.38\linewidth}
    \includegraphics[width=\linewidth]{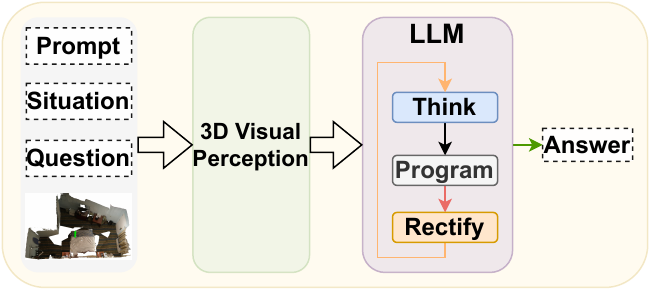}
    \caption{Ours: LLM-TPC.}
    \label{fig:llm_tpc}
    \end{subfigure}
    \vspace{+5pt}
   \caption{Existing methods for 3D-SR task. End-to-end methods lack interpretability and cannot accomplish the 3D-SR task in a zero-shot or few-shot way. Language-only methods fail to conduct multi-modal reasoning and deliver reasonable results. }
\label{fig:existing_methods}
\end{figure*}

Prevalent approaches for 3D-SR mainly adopt end-to-end models (\cref{fig:end_to_end}) \cite{scanqa,sqa3d,3d-vista,3dllm} trained on supervised data to align 3D scenes and texts for answer prediction.
However, it is prohibitively expensive to collect extensive in-domain data that covers sufficient world knowledge and skills to answer questions in-the-wild.
Moreover, these models generate the final answers without an explicit reasoning process, making them less interpretable to humans.
Inspired by the recent success of large language models (LLMs) in solving complex tasks by in-context few-shot learning~\cite{cot,rethinking,leasttomost}, there has been a growing focus on harnessing LLMs for visual reasoning~\cite{Visprog,vipergpt}.
One typical approach (\cref{fig:llm_caption}) is to convert the visual input to texts through visual captioning, and then rely on LLMs to solve the problem~\cite{visualchatgpt,hugginggpt}.
Ma \etal~\cite{sqa3d} makes the first attempt to leverage a powerful LLM GPT-3~\cite{gpt3} for 3D-SR. 
However, it fails to deliver satisfactory performance due to two main reasons.
First, generating an exhaustive caption of the 3D environment is difficult and might not be necessary for the target question.
Second, while LLMs excel in common-sense reasoning, they are limited in performing calculations and symbolic reasoning as demonstrated in~\cite{pal,bbh} such as handling 3D spatial relations.
To alleviate these limitations, the second type of approaches~\cite{binder,pal,Visprog,vipergpt} employ LLMs to write executable programs that decompose the reasoning problem into sub-problems and use appropriate modules to solve each sub-problem.
However, the design of task decomposition and module implementation for the complex 3D-SR task remains an open issue.
Furthermore, existing frameworks only produce programs in a single iteration, leading to a failure rate exceeding 20\% in complex 3D scenarios as shown in our empirical experiments in \cref{tab:ablation_par}. %

In this work, we propose a novel framework \textbf{LLM-TPC} (\cref{fig:llm_tpc}) to better leverage LLMs in solving complex 3D-SR tasks without any training.
It contains a \textbf{T}hink-\textbf{P}rogram-re\textbf{C}tify loop to iteratively enhance the question answering performance.
In the \textbf{\textit{Think}} phase, an LLM is prompted to decompose the question into a series of steps in natural language, taking advantage of LLM's world knowledge.
It then generates an executable Python program in the following \textbf{\textit{Program}} phase guided by the steps in the Think phase. The program calls a set of 3D visual perception modules to query necessary information needed to solve the target question.
We carefully design the 3D visual perception modules for recognizing objects, attributes and 3D spatial relations.
Next, in the \textbf{\textit{Rectify}} phase, the program is executed and corrected if it fails or reaches a maximum number of iterations.
Finally, the final answer is formalized through summarizing the execution results. 

{We conduct extensive experiments on the 3D-SR benchmark SQA3D~\cite{sqa3d}.
The ablation studies demonstrate the effectiveness of the proposed Think-Program-reCtify loop, which can enhance the interpretability and robustness to generate executable codes and thus improves the performance.
The results also highlight the importance of flexible and accurate 3D perception modules. Our automatic attribute and relation recognition modules perform better than incomplete groundtruth captions, while existing 3D object segmentation remains a bottleneck.
When using groundtruth object segmentation labels, our LLM-TPC significantly outperforms existing end-to-end models~\cite{sqa3d,3d-vista,3dllm,leo} and LLM-based baselines. 
Although the automatic object segmentation deteriorates the performance of our model, we find LLM-TPC is still complementary with end-to-end models, where an automatic ensemble establishes a new state-of-the-art (SoTA) on the SQA3D benchmark with and without groundtruth object segmentation.
To gain more insights on the strength of LLM-TPC, we manually divide a subset of SQA3D testing examples according to different required skills such as common-sense and relation understanding, and observe that LLM-TPC excels in answering knowledge-dependent and reasoning-intensive questions. 
}

In summary, the main contributions of this work are as follows:

\parskip=0.1em
\begin{itemize}[itemsep=0.1em,parsep=0em,topsep=0em,partopsep=0em]
    \item[$\bullet$]  We propose LLM-TPC, a training-free framework that leverages large language models and decouples 3D visual perception and reasoning skills for 3D Situated Reasoning.

    \item[$\bullet$]  We design the Think-Program-reCtify loop to jointly integrate the 3D visual perception ability and LLM's excellent reasoning ability through Think and Program, and further enhance robustness through reCtify with iterative self-reflection.%
    \item[$\bullet$]  We conduct extensive experiments and analysis on the SQA3D benchmark and achieve state-of-the-art performance with no need of training, which improves the effectiveness, interpretability, and robustness of 3D-SR.
\end{itemize}

\section{Related Work}
\label{sec:Related Work}

\subsection{3D Scene Understanding}
3D Scene Understanding aims to grasp the semantic meaning of objects and the environment from point clouds. It includes tasks such as object recognition~\cite{objectnet3d, pointformer}, semantic/instance segmentation~\cite{mask3d, 3d-bonet}, grounding~\cite{ns3d, referit3d}, and 3D-VQA (visual question answering)~\cite{scanqa, clevr3d}.
These tasks only demand several certain skills such as spatial reasoning, visual perception, and linguistic understanding. In addition to these basic skills, 3D-SR further requires multiple reasoning skills, including common-sense reasoning and calculation. Such comprehensive skill set requirement makes 3D-SR uniquely much more challenging.

Among all tasks, 3D-VQA is most similar to our task.  Besides the basic QA task~\cite{3dqa}, researchers have been seeking enhanced interpretability with an additional localization task~\cite{scanqa, FE-3DGVQA}, and to address co-occurrence biases~\cite{clevr3d, scanqa, FE-3DGVQA}. However, they often assume a third-person perspective when observing 3D scenes. SQA3D~\cite{sqa3d} stands out by adopting an egocentric view, in hopes of equipping the agent with embodied scene understanding ability. Notably, in contrast to embodiedQA~\cite{EQA, MP3D-EQA}, SQA3D simplifies the protocol to QA-only, resulting in more complex and knowledge-intensive questions. 
Existing end-to-end methods address the task by designing task-specific modules and auxiliary losses  (\eg 3D object detection/classification and text classification~\cite{scanqa, sqa3d}), or by conducting 3D vision and text alignment through pre-training~\cite{3d-vista}. However, the lack of comprehensive 3D vision-text data covering enough world knowledge for 3D-SR limits their effectiveness and interpretability, which inspired us to leverage the commonsense and reasoning ability of LLMs.

\subsection{LLMs for Vision Tasks}

LLMs such as ChatGPT~\cite{chatgpt} have revolutionized Natural Language Processing (NLP), leading researchers to explore their applications in vision tasks. Approaches incorporating other modalities into LLMs can be categorized into two folds: (1) end-to-end training with Large Multi-modal Models (LMMs) through instruction tuning, and 
(2) augmented systems that equip Large Language Models (LLMs) with external tools.
LMM-based methods~\cite{MiniGPT-4, LLaVa, InstructBLIP} convert visual input into LLMs through pretraining for visual-text alignment and downstream tuning for instruction following. 3D-LLM~\cite{3dllm} and LEO~\cite{leo} employ LLMs to gather extensive 3D-language data for pretraining and instruction tuning, convert visual inputs (\eg 2D images and 3D point clouds) into language models' tokens through modality-specific encoders~\cite{clip, PointNet++} to ground the language models into 3D worlds.
Augmented systems~\cite{ProgPrompt, visualchatgpt, MM-REACT, Visprog, vipergpt} leverage LLMs' in-context learning ability to incorporate external tools for visual information. VisProg~\cite{Visprog} and ViperGPT~\cite{vipergpt} address complex 2D visual tasks by generating programs through LLMs and executing them with external tools such as visual foundation models. As for 3D vision tasks, current works~\cite{llm-grounder, zero-grounding, ns3d} mainly focus on addressing 3D visual grounding task by decomposing language queries into sub-queries through LLMs, leaving the complex 3D-SR task an open issue.

\section{Method}
The 3D Situated Reasoning (3D-SR) task aims to answer a question based on the agent's egocentric situation in a 3D environment. 
The input consists of 3D point clouds, textual situation description plus question, and the output is the text answer. 
In this Section, we first present the overall framework of our LLM-TPC in \cref{subsec:overall_framework}, and then introduce 3D visual perception modules in \cref{subsec:api_def} which provide APIs for LLMs. Finally, we describe the reasoning procedure of LLM-TPC in detail in \cref{subsec:par}.

\subsection{Overall Framework}
\label{subsec:overall_framework}

\begin{figure}
    \centering
    \vspace{-10pt}
    \includegraphics[width=\linewidth]{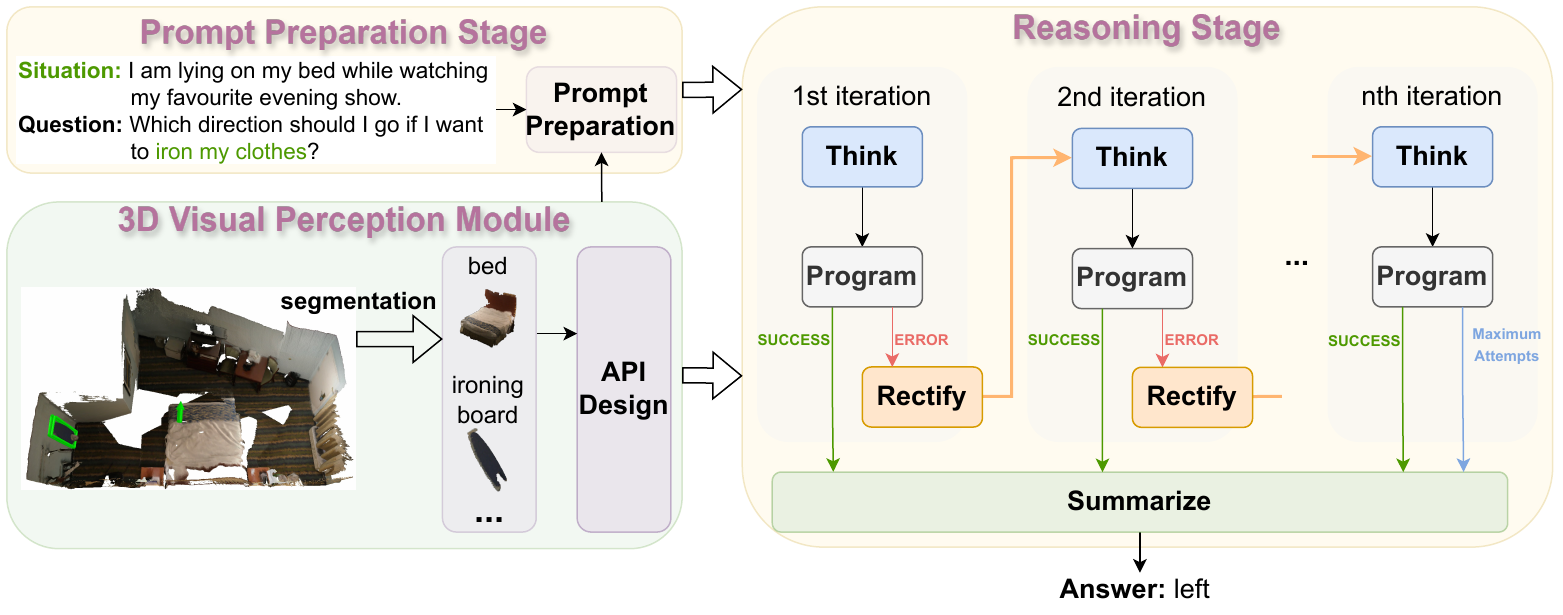}
\vspace{-20pt}
\caption{Overall Framework of LLM-TPC. LLM-TPC comprises three key components: the 3D Visual Perception Module equips the LLM with 3D context perception abilities, the Prompt Preparation Stage prepares prompts for reasoning, and the Reasoning Stage involves iterative Think-Program-reCtify loops.}
    \label{fig:overall_framework}
    \vspace{-10pt}
\end{figure}

\cref{fig:overall_framework} presents the overall framework of our proposed LLM-TPC, which decouples 3D visual perception and reasoning processes in the 3D-SR task.

\noindent
\textbf{3D Visual Perception Module.} It aims to equip the LLM with abilities to perceive the 3D information at different levels of granularity.
It applies segmentation on the raw 3D scan to create an object-level representation of the scene, and recognize object categories, attributes and relationships.

\noindent
\textbf{Prompt Preparation Stage.} 
We prompt the LLM by demonstrating the task definition, format specification, API documentation, and in-context examples. Then we feed an overall summary of object categories in the scene from the 3D Visual Perception Module along with the situation and question.
Please see the supplementary material for more details.

\noindent
\textbf{Reasoning Stage.} 
The LLM iteratively engages in the Think-Program-reCtify (TPC) loop through generating plans, developing programs with API calls to implement the plans, and obtaining answers based on the program's execution results, or adjusting thought and program in case of execution failure.
If the program successfully executes or it reaches the maximum iteration number, it summarizes the final answer.

\subsection{3D Visual Perception Module}
\label{subsec:api_def}

We propose the following 3D visual perception modules for comprehensive 3D scene understanding.

\noindent\textbf{3D object segmentation.}
Given a 3D scan represented by point clouds, we utilize a state-of-the-art model Mask3D~\cite{mask3d} to perform instance segmentation for object-centric representations.
For each segmented instance, we keep its colored point cloud, center position and the 3D bounding box derived from the segmentation mask.

\noindent\textbf{Object category and attribute classification.}
We use OpenShape~\cite{openshape} to predict the category and attributes for each object.
OpenShape~\cite{openshape} is pre-trained on triplets of point clouds, images and texts, and generates point cloud embeddings that are in the same space as the pre-trained CLIP embeddings~\cite{clip}.
Therefore, it can perform zero-shot classification given the point cloud.
Specifically, we feed the object point cloud and text candidates of object categories or attributes into OpenShape. Then, we calculate the similarity between the 3D feature and textual features of each candidate. We choose the one with the highest similarity.
Please see the supplementary material for more details.

\begin{table}[t]
    \small
    \centering
    \caption{API definition in LLM-TPC. They're split into 4 categories: Scene Description (SD), Object Filtering (OF), Object Querying by Relation (OQR), and Object Information Querying (OIQ).}
    \label{tab:api_def}
    \vspace{-8pt}
    \begin{tabular}{m{0.04\linewidth}>{\arraybackslash}m{0.33\linewidth}>{\arraybackslash}m{0.61\linewidth}}
         \toprule
         & \multicolumn{1}{c}{definition} & \multicolumn{1}{c}{semantics} \\
         \midrule
         \rotatebox[]{90}{\textbf{SD}} & $scene() \longrightarrow y$ & Returns all objects $y$ in the 3D scene, containing their positions, attributes, etc. \\
         \midrule
         \rotatebox[]{90}{\textbf{OF}} & $filter(x, c) \longrightarrow y$ & Returns a set of objects $y$ that belongs to category $c$.\\
         \midrule
         \multirow{3}{*}{\rotatebox[]{90}{\textbf{OQR}}} & 
         $relate(x^t, x^r, r) \longrightarrow y$ & Returns a set of objects $y$ that are related to the reference object $x^r$ by the relation $r$.\\
         & $relate\_agent(x^t, r) \longrightarrow y$ & Returns a set of objects $y$ that are related to the agent by the relation $r$.\\
         \midrule
         \multirow{6}{*}{\rotatebox[]{90}{\textbf{OIQ}}} &
         $query\_relation(x^t, x^r) \longrightarrow rels$ & Returns a list of relations $rels$ between the object $x^t$ and the reference object $x^r$. \\
         & $query\_relation\_agent(x^t)$ $\longrightarrow rels$ & Returns a list of relations $rels$ between the object $x^t$ and the agent. \\
         \cmidrule{2-3}
         & $query\_attribute(x, a\_type$ $ [, a\_cands]) \longrightarrow  val$ & Returns the value $val$ of attribute type $a\_type$ (\eg color, shape, etc.) of the object $x$, or select attribute value from the candidates $a\_cands$ that best matches the characteristics of the object. \\
         \bottomrule
    \end{tabular}
    \vspace{-10pt}
\end{table}

\noindent\textbf{Spatial relation recognition.}
Given two objects, the goal is to recognize their spatial relations of three types, including horizontal (closest, farthest), vertical (above, on, etc.), and allocentric (left, right, etc.). By comparing the objects' distance, IOU, and occupancy rate with the pre-defined parameters, we can derive whether they belong to the given relation.
Details are presented in the supplementary material.

Based on these 3D perception modules, we offer four types of APIs within LLM-TPC, as presented in \cref{tab:api_def}:
(i) Scene Description (SD): Provides comprehensive information about all objects in the scene.
(ii) Object Filtering (OF): Filters input objects based on the specified category.
(iii) Object Querying by Relation (OQR): Identifies objects that satisfy a given relation with the target object or the agent.
(iv) Object Information Querying (OIQ): Retrieves relations or attributes of provided object(s).

\subsection{Reasoning Stage with LLMs}
\label{subsec:par}
\noindent
\textbf{Think and Program.} 

\begin{wrapfigure}{r}{0.6\textwidth}
    \vspace{-40pt}
    \begin{minipage}{0.6\textwidth}
        \centering
        \includegraphics[width=\linewidth]{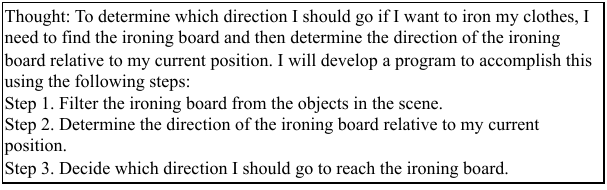}
        \subcaption{Step-by-step plans generated in the Think phase.}
        \label{fig:think-1}
    \end{minipage}
    \begin{minipage}{0.6\textwidth}
        \centering
        \includegraphics[width=\linewidth]{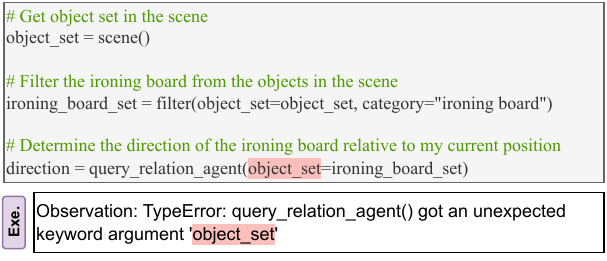}
        \subcaption{Program and execution results in the Program phase.}
        \label{fig:program-1}
    \end{minipage}
    \begin{minipage}{0.6\textwidth}
        \centering
        \includegraphics[width=\linewidth]{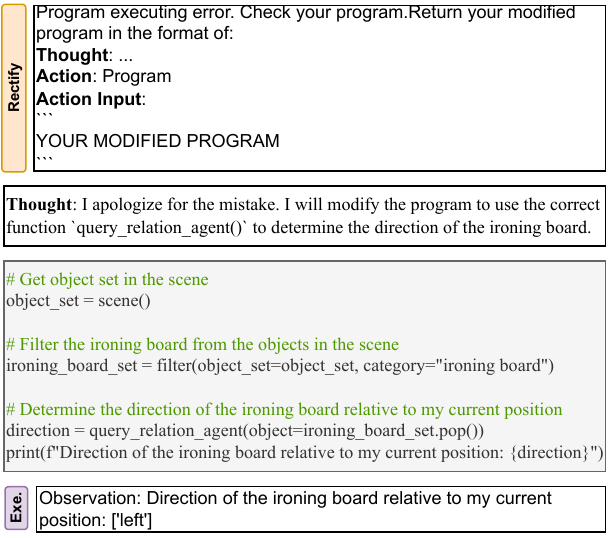}
        \subcaption{Modified plan and program triggered in the Rectify phase.}
        \label{fig:rectify-1}
    \end{minipage}
    \begin{minipage}{0.6\textwidth}
        \centering
        \includegraphics[width=\linewidth]{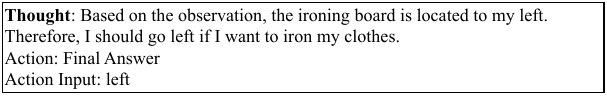}
        \subcaption{Final answer in the Summarization phase.}
        \label{fig:summary-1}
    \end{minipage}
\caption{Example outputs of LLM-TPC}{}
\label{fig:example_TPC}
\vspace{-30pt}
\end{wrapfigure}

\noindent
In the Think phase, LLM-TPC decomposes the compositional question into a sequence of steps (\cref{fig:think-1}).
In the Program phase, it generates an executable Python program to ground each step to a piece of code and extract relevant concepts (e.g., spatial relationships, layout, color, shape, material of objects) from the 3D scene through API calls if necessary (\cref{fig:program-1}). 
Then the program is executed.

\noindent
\textbf{Rectify via self-reflection.}
`Think and Program' in one turn can not guarantee to come up with the perfect plan, which may lead to program errors, thus we further design a Rectify module, which triggers LLM's self-reflection mechanism to adjust the plan and modify the program error. Concretely, when the execution fails, LLM-TPC enters the Rectify module, where the LLM is prompted with a debug command to loop back to the Think-Program phase and adjusts the plan and program (\cref{fig:rectify-1}) according to the received error message.

\noindent
\textbf{Summarization.}
Once the TPC loop completes its iterations and generates a refined program or the maximum number of attempts is reached, LLM-TPC integrates the execution results of each iteration to concludes the final answer, as shown in \cref{fig:summary-1}.
Besides the information derived from the 3D scene, common-sense knowledge is also important for the 3D-SR task because it helps to infer object proximity, functionality, and potential actions of individuals within specific contexts. 
Leveraging common-sense reasoning in the summarization phase can lead to knowledge-intensive answers that go beyond the information derived solely from the 3D scene. Examples can be found in the supplementary material.

\noindent\textbf{Ensemble.}
{
As our LLM-TPC is training-free, it ignores domain specific knowledge such as answer distribution priors compared to end-to-end models, while end-to-end models are less generalizable and lack complex reasoning abilities.
Therefore, combining results from LLM-TPC and prior end-to-end models can take the best of both worlds.
To this end, we further propose to use an LLM to automatically ensemble results from our LLM-TPC and prior end-to-end models.
Specifically, we merge the top 5 answers and probabilities from an end-to-end model and the open-ended responses from our LLM-TPC, and feed them into another LLM to generate the final answer. The detailed prompt for the ensemble is presented in the supplementary material.
}

\section{Experiment}
\subsection{Experimental Setup}
\noindent
\textbf{Dataset.}
We evaluate on the SQA3D~\cite{sqa3d} benchmark, the current largest dataset for grounded 3D scene understanding. It is more challenging than other 3D-QA tasks~\cite{scanqa,clevr3d} as it requires multiple reasoning skills such as common-sense, calculation, multi-hop reasoning, \etc.
SQA3D comprises 20.4k descriptions of 6.8k unique situations collected from 650 ScanNet \cite{scannet} scenes, along with 33.4k questions about these situations. 
The question-answering pairs are divided into 26,623 training samples, 3,261 validation samples, and 3,519 testing samples. 
As our method is training-free, we only select three samples from the training set as in-context examples for the LLM.
We evaluate all the approaches on the full testing set.

\noindent\textbf{Implementation details.}
We use ChatGPT \cite{chatgpt} in our LLM-TPC via its \verb+gpt-3.5-turbo-16k+ API and three examples for in-context learning.
Details can be found in the supplementary material.

\noindent\textbf{Compared methods.}
We compare with both end-to-end supervised methods, including ScanQA~\cite{sqa3d}, 3D-VisTA~\cite{3d-vista}, 3D-LLM~\cite{3dllm} and LEO~\cite{leo}, that are trained on the SQA3D training set, and LLM-based baseline LLM-T, which is a variant of our LLM-TPC.
Please see the supplementary material for more details.

\begin{itemize}
    \item ScanQA~\cite{sqa3d}: In the setting without groundtruth (w/o GT), ScanQA processes the entire 3D scan and extracts object-centric features using VoteNet~\cite{votenet}. In the setting with groundtruth (w/ GT), ScanQA is provided with groundtruth object labels, RGB colors, location, and size features for retraining.
    \item 3D-VisTA~\cite{3d-vista}: It first applies segmentation masks on the whole 3D scan to extract object point clouds and then obtain their object features for multi-modal fusion with text tokens. We apply segmentation masks predicted by Mask3D~\cite{mask3d} and groundtruth segmentation masks in the setting w/o and w/ GT, respectively.
    \item 3D-LLM~\cite{3dllm}: It first extracts 2D dense features from multi-view images and then map 2D features into reconstructed 3D points. It uses pretrained 2D VLM as backbone and inputs the aligned 3D features to train the 3D-LLM.
    \item LEO~\cite{leo}: It first applies segmentation masks on the whole 3D scan to extract object point clouds and then send multi-modal tokens including text tokens, 2D tokens encoded from egocentric image and 3D object tokens to the pretrained LLM for prefix language modeling. We apply segmentation masks predicted by Mask3D~\cite{mask3d} and groundtruth segmentation masks in the setting w/o and w/ GT, respectively.
    \item LLM-T is the Think-only variant of LLM-TPC. It uses 30 groundtruth captions from ScanRefer~\cite{scanrefer} dataset as the visual context. The captions are concatenated with situation description and question as the prompt to ChatGPT for answer generation. Three training examples are used for in-context learning, and is prepended with additional task instruction to restrict the generation of answers with no more than 3 words. 
\end{itemize}

\noindent\textbf{Evaluation protocol.}
SQA3D~\cite{sqa3d} adopts "\emph{strict match}" as the evaluation metric, \ie the accuracy of answer classification given 706 candidate answers in the test set.
However, this evaluation metric is not well-suited for assessing open-ended answers generated by LLMs. 
For instance, given the question "Which room has more light, one in front of me or one behind me?", the response generated by the LLM "in front of me" is correct although it does not exactly match the groundtruth answer "front". 
Therefore, we adopt a "\emph{soft match}" metric to evaluate LLM-generated answers: if the predicted answer has an intersection with the groundtruth words, or if they comply with our predefined synonym rules, the prediction is deemed as correct. Details can be found in the supplementary material.

\subsection{Comparison with State of the Arts}

\begin{table}[h]
\centering
\caption{Answer accuracy (\%) on the full test set of SQA3D. "GT" denotes the groundtruth information.}
\label{tab:compare_end_to_end}
\vspace{-8pt}
\tabcolsep=0.25cm
\begin{tabular}{ll|cc|c} \toprule
 & \multirow{2}{*}{Method} & \multicolumn{2}{c|}{w/o GT}  & \multicolumn{1}{c}{w/ GT}\\
\cmidrule(r){3-5} 
 & & Proposal & Acc  &  Acc \\ \midrule
\multirow{4}{*}{Supervised} & ScanQA~\cite{sqa3d} & VoteNet~\cite{votenet} & 46.38  & 47.74\\
 & 3D-VisTA~\cite{3d-vista} & Mask3D~\cite{mask3d} & 50.44  & 50.72 \\
 & 3D-LLM~\cite{3dllm} & - & 50.21  & - \\
 & LEO~\cite{leo} & Mask3D~\cite{mask3d} & 52.83  & 53.25 \\ \midrule
\multirow{2}{*}{Few-shot} & LLM-T & - & -  & 48.28 \\
 & LLM-TPC (Ours) & Mask3D~\cite{mask3d} & 45.84  & 56.92 \\ \midrule
Ensemble & LLM-TPC + 3D-VisTA & Mask3D~\cite{mask3d} & \textbf{55.81}  & \textbf{58.20} \\ \bottomrule
\end{tabular}
\vspace{-15pt}
\end{table}

\cref{tab:compare_end_to_end} compares our approach with state-of-the-art methods on SQA3D dataset. 
We conduct experiments in two settings, with or without groundtruth information in 3D visual perception. 
When using groundtruth information, our proposed LLM-TPC outperforms both end-to-end supervised methods and its Think-only variant LLM-T, achieving significant gains of $+3.67\%$ and $+8.64\%$, respectively.
In settings without groundtruth information, we apply VoteNet predicted proposals for ScanQA, and Mask3D predicted segmentation masks covering 200 categories for 3D-VisTA, LEO and LLM-TPC. In this setting, LLM-TPC underperforms end-to-end methods due to limitations of its 3D visual perception, which will be discussed in \cref{subsec:ablation}.%

Interestingly, for the three end-to-end models, using groundtruth information only brings trivial improvement, which suggests that visual perception might not be the main bottleneck for end-to-end models, and insufficient reasoning capabilities may be their main drawback, since the 3D-SR task requires a variety of reasoning skills including common-sense reasoning. %

{The predictions of LLM-TPC and end-to-end method are complementary with each other as shown in the last row of \cref{tab:compare_end_to_end}. 
The proposed ensemble with LLM further enhances the performance in both settings with and without groundtruth information (1.28\% and 9.97\% gains respectively), establishing a new state of the art on the benchmark.}

\subsection{Ablation Study}
\label{subsec:ablation}

\begin{table}[h]
    \centering
    \vspace{-20pt}
    \caption{Ablation on different 3D visual perception modules. We evaluate the answer accuracy (\%) on a subset of 800 samples from the test set. "GT" denotes the groundtruth.}
    \vspace{-8pt}
    \label{tab:ablation_context}
    \tabcolsep=0.35cm
    \begin{tabular}{c|cccc} \toprule
           & Object seg & Object label & Attribute & Acc \\
            \midrule
            r1 &GT & GT & OpenShape & 57.50 \\
            r2 & GT & GT & caption & 52.00 \\
            r3 & GT & OpenShape & OpenShape & 50.00 \\
            r4 & Mask3D & OpenShape & OpenShape & 46.00 \\
            r5 & Mask3D & Mask3D & OpenShape & 48.12 \\
            \bottomrule
    \end{tabular}
    \vspace{-20pt}
\end{table}

\noindent
\textbf{3D visual perception.}
\cref{tab:ablation_context} compares the influence of using different 3D visual perception modules to provide visual context for LLM-TPC, including the object location, category and attributes.
For the object attributes, we compare the ones predicted by OpenShape~\cite{openshape} and the ones extracted from the groundtruth captions.%
We can see from r1 and r2 in \cref{tab:ablation_context} that the automatically predicted attributes perform much better than the extracted attributes from the groundtruth captions, leading to $5.5\%$ difference in accuracy. This is because the groundtruth captions cannot exhaustively list all the required attributes asked by the LLM's program, while the neural network based method OpenShape is more flexible to satisfy the calls from the LLM.
For the object category, we also compare the groundtruth labels and the ones predicted by OpenShape in r1 and r3. The performance drops by 7.5\% with the automatic predictions, because the incorrect predictions confuse the LLM, leading to wrong answers.
Another significant performance decrease is from automatic object segmentation in r4 and r5.
Though current automatic 3D perception modules limit the reasoning ability of LLMs, we believe the modular design of our method makes it easier to benefit from the advancement in the 3D field.

\begin{wraptable}{r}{0.6\textwidth}
    \vspace{-20pt}
    \centering
    \footnotesize
    \caption{Ablations on the effectiveness of the Think, Program and Rectify modules.
    Results are presented in answer accuracy (\%), program pass rate (\%) and the average number of interaction rounds on a subset of 800 samples from the test set.}
    \label{tab:ablation_par}
    \begin{tabular}{cccccc}
        \toprule
        Setting & Acc & Pass rate & Rounds \\
        \midrule
        \multicolumn{1}{l}{Think} & 46.75 & - & 1 \\
        \multicolumn{1}{l}{Think-Program} & 50.25 & 79.87 & 1.97 \\
        \multicolumn{1}{l}{Think-Program-Rectify} & 57.50 & 96.93 & 2.35 \\
        \bottomrule
    \end{tabular}
    \vspace{-15pt}
\end{wraptable}

\noindent\textbf{Think-Program-reCtify loop.}
\cref{tab:ablation_par} validates the effectiveness of each part of our TPC loop. We report the answer accuracy, program pass rate, and average number of interaction rounds. 
The first row is LLM-T which only uses natural language reasoning to generate answers.
It performs much worse than the other variants, indicating difficulty in addressing 3D-SR task using plain LLMs.
Integrating Program with Think improves the accuracy by 3.5\% as it allows to leverage 3D contexts provided by 3D visual perception module through programming.
However, it still suffers from the relatively low program passing rate. 
The Rectify step triggers LLM's self-reflective mechanism with increased interaction rounds, significantly improving the success rate of programming from 79.87\% to 96.93\%.%

\noindent
\textbf{Influence of maximum iteration number.}

\begin{wrapfigure}{r}{0.4\textwidth}
    \vspace{-35pt}
    \includegraphics[width=\linewidth]{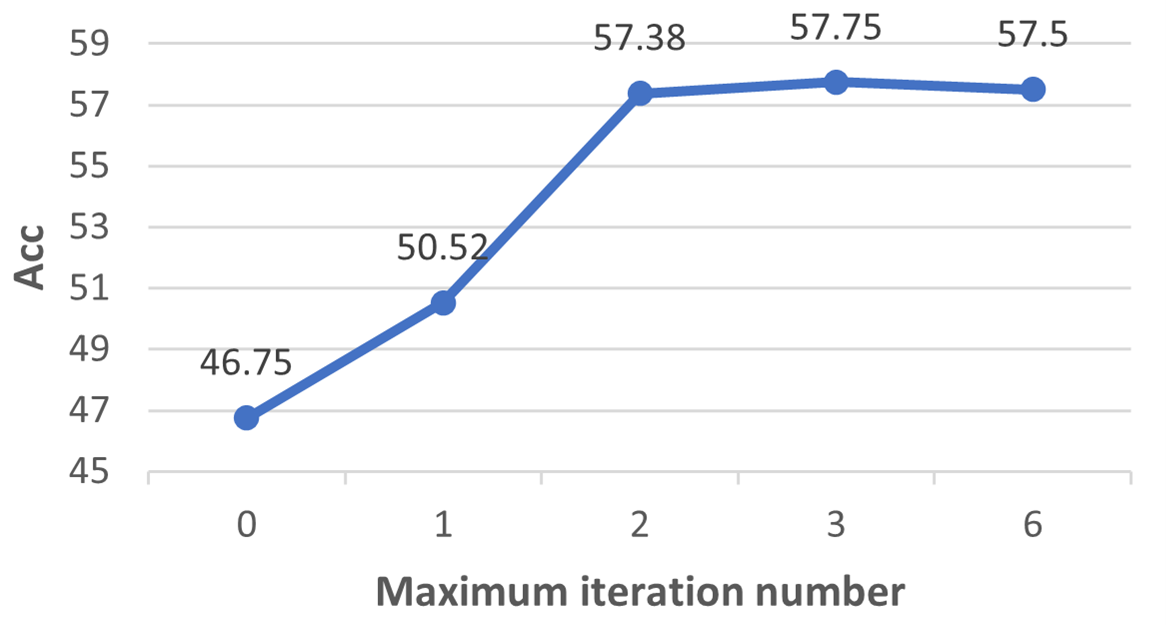}
\caption{Influence of maximum iteration number. Results are presented in answer accuracy (\%) on a subset of 800 samples from the test set.}
\label{fig:max_iteration_num}
\vspace{-25pt}
\end{wrapfigure}

\noindent
\cref{fig:max_iteration_num} presents results with different maximum iteration numbers. The performance continues to improve with more iterations and reaches convergence after 2 iterations. This is because 95.9\% of the samples can be covered within 3 iterations.
Further increasing the maximum iteration number does not bring performance gain. 
This can be attributed to the limitations of the current "Rectify" module, which can only provide feedback on program errors and formatting issues with limited suggestions for correction. When the LLM fails to correct errors after multiple attempts, excessive historical information can lead to catastrophic forgetting, resulting in decreased performance. By designing a more informative correction mechanism to provide clearer feedback, LLM-TPC's self-correction capability can be enhanced.

\subsection{Further Breakdown Analysis}

\begin{table}[h]
    \centering
    \vspace{-8pt}
    \caption{Breakdown analysis of accuracy (\%) according to different question types. The evaluation is conducted on a subset of 800 samples from SQA3D test set. We compare 3D-VisTA~\cite{3d-vista}, three variants of our LLM-TPC model where T denotes Think only, TP combines Think and Program, and TPC further performs Rectify loop, along with the ensemble approach.}
    \label{tab:break_down}
    \vspace{-8pt}
    \tabcolsep=0.25cm
    \begin{tabular}{cccccccc}
        \toprule
            \multirow{2}{*}{Question type} & \multirow{2}{*}{Num} & \multicolumn{5}{c}{Method}\\
        \cline{3-7}
           & & 3D-VisTA & T & TP & TPC & Ensemble \\
        \midrule
            common-sense & 104 & 57.69 & 51.92 & 53.85 & 62.50 & \textbf{73.08} \\
             embodied activity & 133 & 60.90 & 45.11 & 55.64 & \textbf{68.42} & 66.92 \\
             navigation & 135 & 40.74 & 31.11 & 57.04 & \textbf{77.04} & 57.78 \\
             multi-hop reasoning & 285 & 51.58 & 41.05 & 50.88 & 59.30 & \textbf{62.81} \\
             calculation & 216 & 54.63 & 44.44 & 60.19 & \textbf{65.28} & 60.19 \\
             relation & 756 & 40.49 & 43.90 & 30.73 & 39.02 & \textbf{53.17} \\
             visual concepts & 192 & 55.21 & 57.29 & 38.54 & 39.06 & \textbf{59.90} \\
             \midrule
             Acc & 800 & 51.00 & 46.75 & 50.25 & 57.50 & \textbf{59.62} \\
        \bottomrule
    \end{tabular}
    \vspace{-15pt}
\end{table}

To gain a more in-depth understanding of our method's performance, we manually classify questions into seven types according to the required reasoning skills. A question can fall into multiple categories if it requires multiple skills.
The seven categories include:
(i) common-sense: involving general knowledge and reasoning that is commonly understood and expected;
(ii) embodied activity: These questions pertain to activities or actions carried out by an embodied agent within the 3D environment.
(iii) navigation: Questions related to spatial orientation, movement, and navigation within the 3D scene.
(iv) multi-hop reasoning: Questions requiring sequential reasoning across multiple ($\geq 3$) steps to arrive at the answer.
(v) relation: Questions involving spatial relationships between objects and/or the agent.
(vi) calculation: Questions requiring mathematical operations, calculations, and measuring distances within the 3D environment.
(vii) visual concepts: Questions about the color, shapes, size, material, and other attributes that require visual perception of objects in the scene.
We annotate 800 questions from the 3519 questions in the SQA3D test set.

We conduct a comparison experiment with the end-to-end method (3D-VisTA~\cite{3d-vista}) and different variants of our LLM-TPC in \cref{tab:break_down}. 
Due to its ability to decouple visual perception and reasoning, LLM-TPC demonstrates outstanding performance on reasoning-based questions. 
The plain Think alone (T) is insufficient to address complex reasoning process.
Combining it with Program (TP) grounds each step to a piece of code with API calling, enabling itself to obtain scene information and excel in solving computation-related problems. 
However, LLM-TPC performs poorly on questions about relation and visual concepts. 

To diagnose the cause of failure related to visual concepts (whether due to LLM's reasoning errors or visual recognition errors), we manually examine failure cases related to visual concepts in \cref{tab:break_down}. Among the 117 failure samples, 34.19\% are attributed to LLM's reasoning errors. After manually correcting these instances, we run the 117 groundtruth programs and find that 94.02\% of failures stem from visual recognition errors (\eg incorrect attributes recognized by OpenShape), highlighting the potential for improvement by enhancing visual foundation models.

\begin{figure}[t]
  \centering
    \includegraphics[width=0.9\linewidth]{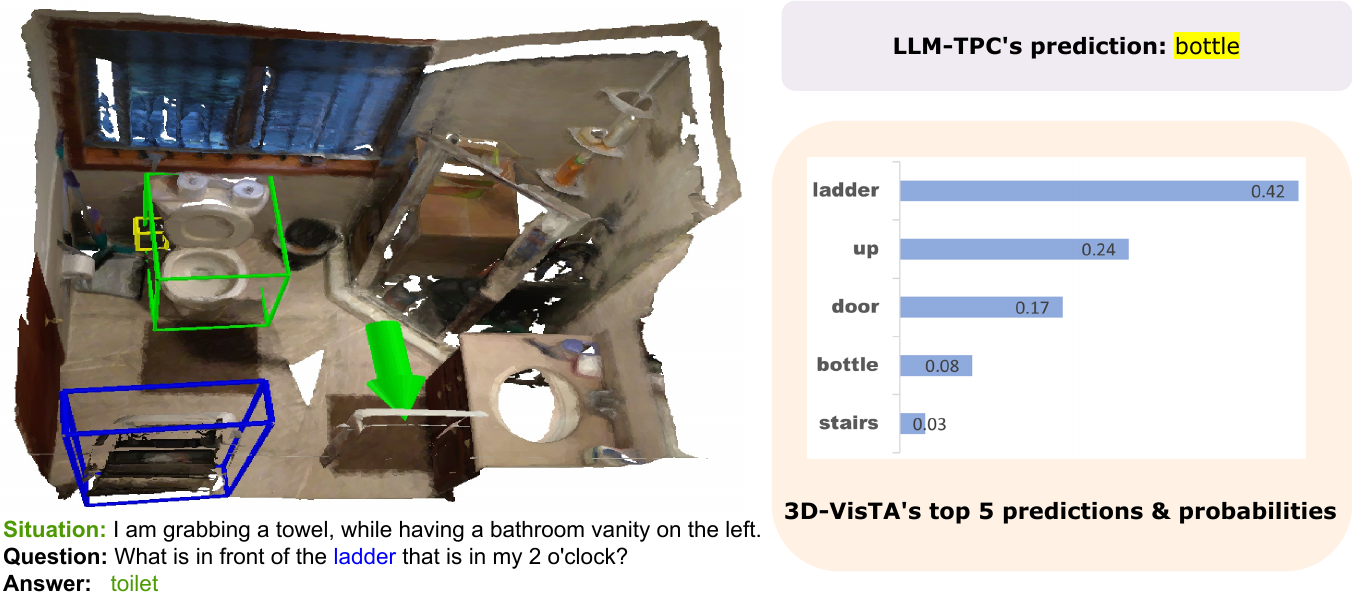}
    \vspace{-10pt}
   \caption{Insufficient answer annotation in SQA3D. Multiple objects (\eg \textcolor{dark-green}{toilet}, door, \textcolor{yellow}{bottle}, window, toilet paper, \etc) are in front of the \textcolor{blue}{ladder}. But only one groundtruth answer is provided.}
   \label{fig:top5}
   \vspace{-15pt}
\end{figure}

Regarding relation-related failures, both 3D-VisTA and LLM-TPC struggle with such questions due to issues with the SQA3D dataset, where a question might have multiple correct answers, but only one correct answer is annotated as shown in \cref{fig:top5}. This ambiguity leads to confusion in 3D-VisTA's predictions, resulting in low confidence for each predicted answer. Additionally, the correct answer predicted by LLM-TPC is also incorrectly evaluated as wrong. The ensemble of LLM-TPC and 3D-VisTA mitigates this issue by incorporating 3D-VisTA's top 5 predictions, leading to a notable enhancement in addressing relation-based questions, which account for 94.5\% of the ablation sample set.

\subsection{Qualitative Results}

\begin{figure}[t]
    \vspace{-8pt}
    \centering
    \begin{subfigure}{\linewidth}
        \centering
        \includegraphics[width=\linewidth]{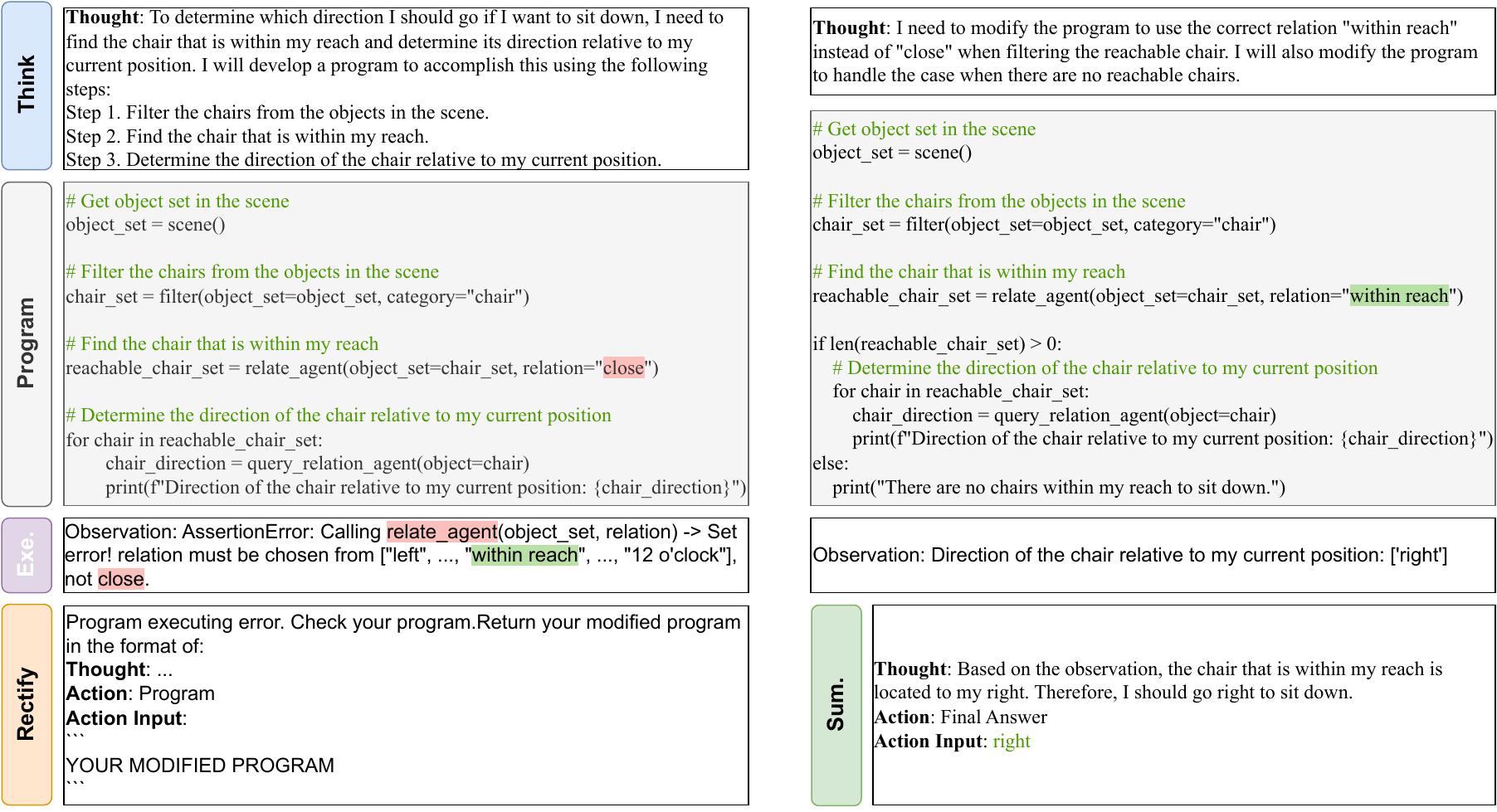}
        \caption{The Rectify module triggers LLM's self-reflection mechanism based on the error information to adjust the plan and modify the program.}
        \label{fig:case_ref}
    \end{subfigure}
    \begin{subfigure}{0.5\linewidth}
        \centering
        \includegraphics[width=\linewidth]{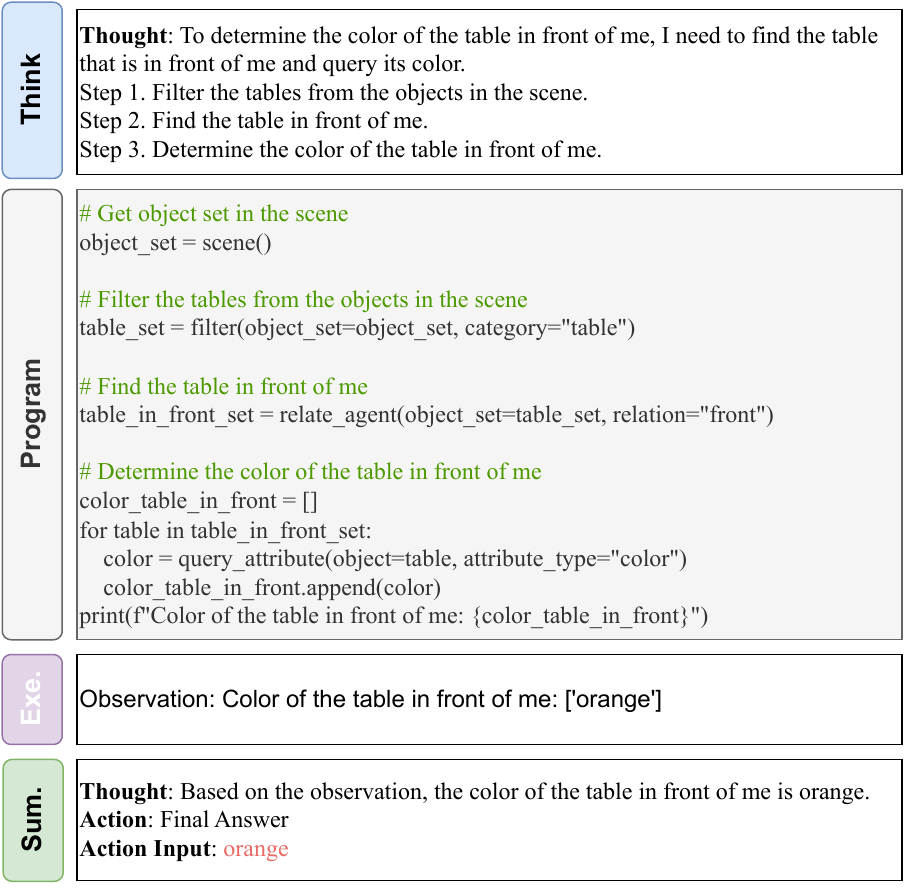}
        \caption{Failure case. Due to the limitations of the API, LLM-TPC mistakenly classified the brown table as orange.}
        \label{fig:case_fail}
    \end{subfigure}
    \hfill
    \begin{subfigure}{0.45\linewidth}
        \centering
        \includegraphics[width=\linewidth]{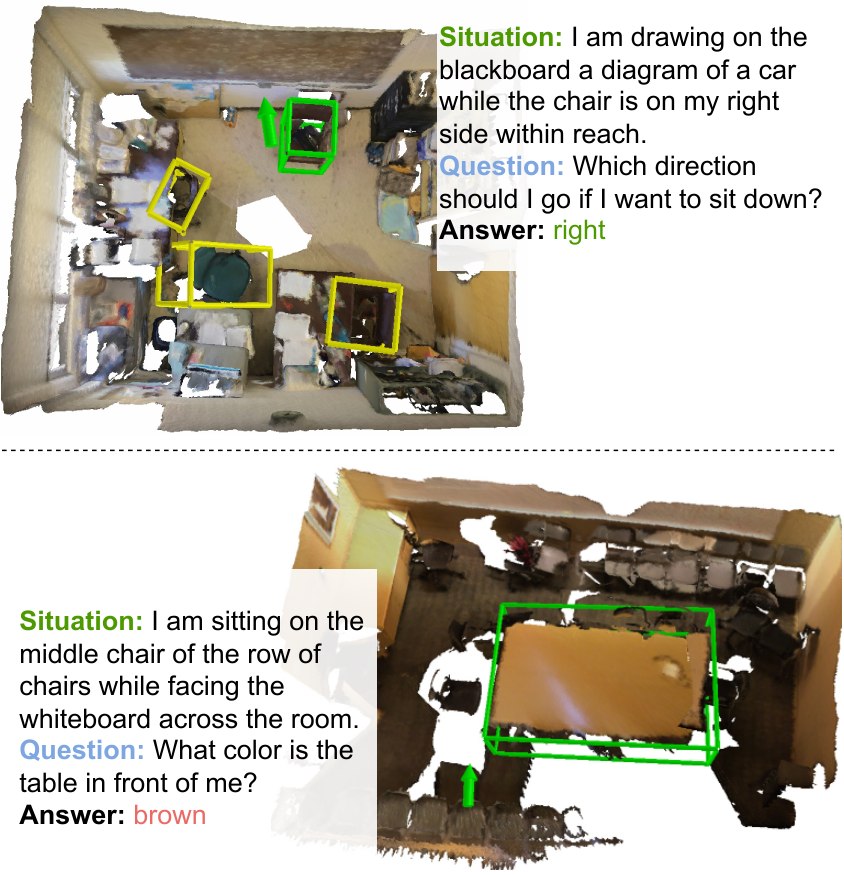}
        \caption{3D scene and situated question answer for \cref{fig:case_ref} (top) and \cref{fig:case_fail} (bottom).}
        \label{fig:case_scene}
    \end{subfigure}
    \vspace{-10pt}
\caption{Qualitative results of LLM-TPC.}
\label{fig:case}
\vspace{-25pt}
\end{figure}

LLM-TPC exhibits strong interpretability, allowing us to easily grasp the reasoning process behind its answers. As illustrated in \cref{fig:case_ref}, LLM-TPC initially formulates step-by-step plans in natural language, then generates programs to implement these plans, and derives the final answer based on the program execution results. By running programs generated by LLM-TPC, we can ground the objects semantically related to the question into the 3D scene. Failure case (\cref{fig:case_fail}) shows that due to the limitations of the API, LLM-TPC mistakenly
classified the brown table as orange.

Since we can detect the capabilities of APIs and foundational visual models by observing the intermediate output of program execution, performance of LLM-TPC can be improved by utilizing more robust APIs and advanced foundational visual models.

\section{Conclusion}
In this work, we propose LLM-TPC for the task of 3D Situated Reasoning (3D-SR). LLM-TPC decouples 3D visual perception from reasoning by employing the Think-Program-reCtify loop. It acquires 3D environmental information via program-based API calling and conducts reasoning using interleaved format of natural language and program. The Rectify mechanism enables LLM-TPC to self-correct its reasoning process. Experiments on the SQA3D benchmark demonstrate the effectiveness, interpretability and robustness of our method.

\noindent\textbf{Limitation.}
LLM-TPC's efficacy is highly related to the accuracy of instance segmentation and attribute classification, and the quality of the APIs it engages with. The current visual perception models with limited ability often leads to incorrect object segmentation and attribute identification, thereby introducing errors into LLM's reasoning process. Furthermore, the output of LLM may deviate from the provided instructions.

While our method serves as a versatile framework encompassing skills from visual perception, spatial reasoning, and more to address various 3D understanding challenges, it remains unexplored in various tasks. Additional experiments on diverse tasks should be conducted in future investigations to further verify the applicability of our framework.

\bibliographystyle{splncs04}

\begin{thebibliography}{10}
\providecommand{\url}[1]{\texttt{#1}}
\providecommand{\urlprefix}{URL }
\providecommand{\doi}[1]{https://doi.org/#1}

\bibitem{referit3d}
Achlioptas, P., Abdelreheem, A., Xia, F., Elhoseiny, M., Guibas, L.: Referit3d: Neural listeners for fine-grained 3d object identification in real-world scenes. In: ECCV (2020)

\bibitem{scanqa}
Azuma, D., Miyanishi, T., Kurita, S., Kawanabe, M.: Scanqa: 3d question answering for spatial scene understanding. In: CVPR (2022)

\bibitem{gpt3}
Brown, T., Mann, B., Ryder, N., Subbiah, M., Kaplan, J.D., Dhariwal, P., Neelakantan, A., Shyam, P., Sastry, G., Askell, A., et~al.: Language models are few-shot learners. NeurIPS  \textbf{33} (2020)

\bibitem{scanrefer}
Chen, D.Z., Chang, A.X., Nie{\ss}ner, M.: Scanrefer: 3d object localization in rgb-d scans using natural language. In: ECCV (2020)

\bibitem{vil3dref}
Chen, S., Guhur, P.L., Tapaswi, M., Schmid, C., Laptev, I.: Language conditioned spatial relation reasoning for 3d object grounding. In: NeurIPS (2022)

\bibitem{binder}
Cheng, Z., Xie, T., Shi, P., Li, C., Nadkarni, R., Hu, Y., Xiong, C., Radev, D., Ostendorf, M., Zettlemoyer, L., Smith, N.A., Yu, T.: Binding language models in symbolic languages. ICLR  (2023)

\bibitem{scannet}
Dai, A., Chang, A.X., Savva, M., Halber, M., Funkhouser, T., Nie{\ss}ner, M.: Scannet: Richly-annotated 3d reconstructions of indoor scenes. In: CVPR (2017)

\bibitem{InstructBLIP}
Dai, W., Li, J., Li, D., Tiong, A.M.H., Zhao, J., Wang, W., Li, B., Fung, P., Hoi, S.: {{InstructBLIP}}: {{Towards General-purpose Vision-Language Models}} with {{Instruction Tuning}} (2023)

\bibitem{EQA}
Das, A., Datta, S., Gkioxari, G., Lee, S., Parikh, D., Batra, D.: Embodied question answering. In: CVPR (2018)

\bibitem{pal}
Gao, L., Madaan, A., Zhou, S., Alon, U., Liu, P., Yang, Y., Callan, J., Neubig, G.: Pal: Program-aided language models. In: ICML (2023)

\bibitem{Visprog}
Gupta, T., Kembhavi, A.: Visual programming: Compositional visual reasoning without training. CVPR  (2023)

\bibitem{3dllm}
Hong, Y., Zhen, H., Chen, P., Zheng, S., Du, Y., Chen, Z., Gan, C.: 3d-llm: Injecting the 3d world into large language models. NeurIPS  (2023)

\bibitem{ns3d}
Hsu, J., Mao, J., Wu, J.: Ns3d: Neuro-symbolic grounding of 3d objects and relations. CVPR  (2023)

\bibitem{leo}
Huang, J., Yong, S., Ma, X., Linghu, X., Li, P., Wang, Y., Li, Q., Zhu, S.C., Jia, B., Huang, S.: An embodied generalist agent in 3d world. arXiv preprint arXiv:2311.12871  (2023)

\bibitem{LLaVa}
Liu, H., Li, C., Wu, Q., Lee, Y.J.: Visual {{Instruction Tuning}} (2023)

\bibitem{openshape}
Liu, M., Shi, R., Kuang, K., Zhu, Y., Li, X., Han, S., Cai, H., Porikli, F., Su, H.: Openshape: Scaling up 3d shape representation towards open-world understanding (2023)

\bibitem{sqa3d}
Ma, X., Yong, S., Zheng, Z., Li, Q., Liang, Y., Zhu, S.C., Huang, S.: Sqa3d: Situated question answering in 3d scenes. In: ICLR (2023)

\bibitem{rethinking}
Min, S., Lyu, X., Holtzman, A., Artetxe, M., Lewis, M., Hajishirzi, H., Zettlemoyer, L.: Rethinking the role of demonstrations: What makes in-context learning work? In: EMNLP (2022)

\bibitem{indoorsu}
Naseer, M., Khan, S.H., Porikli, F.M.: Indoor scene understanding in 2.5/3d for autonomous agents: A survey. IEEE Access  \textbf{7},  1859--1887 (2018)

\bibitem{chatgpt}
OpenAI: Introducing chatgpt (2023), \url{https://openai.com/blog/chatgpt}

\bibitem{pointformer}
Pan, X., Xia, Z., Song, S., Li, L.E., Huang, G.: 3d object detection with pointformer. In: CVPR. pp. 7463--7472 (2021)

\bibitem{glove}
Pennington, J., Socher, R., Manning, C.D.: Glove: Global vectors for word representation. In: Proceedings of the 2014 conference on empirical methods in natural language processing (EMNLP). pp. 1532--1543 (2014)

\bibitem{votenet}
Qi, C.R., Litany, O., He, K., Guibas, L.J.: Deep hough voting for 3d object detection in point clouds. In: ICCV (2019)

\bibitem{PointNet++}
Qi, C.R., Yi, L., Su, H., Guibas, L.J.: {{PointNet}}++: {{Deep Hierarchical Feature Learning}} on {{Point Sets}} in a {{Metric Space}}. In: Advances in {{Neural Information Processing Systems}}. vol.~30. Curran Associates, Inc. (2017)

\bibitem{clip}
Radford, A., Kim, J.W., Hallacy, C., Ramesh, A., Goh, G., Agarwal, S., Sastry, G., Askell, A., Mishkin, P., Clark, J., et~al.: Learning transferable visual models from natural language supervision. In: ICML (2021)

\bibitem{ScanNet200}
Rozenberszki, D., Litany, O., Dai, A.: Language-{{Grounded Indoor 3D Semantic Segmentation}} in~the~{{Wild}}. In: Avidan, S., Brostow, G., Ciss{\'e}, M., Farinella, G.M., Hassner, T. (eds.) Computer {{Vision}} -- {{ECCV}} 2022. pp. 125--141. Lecture {{Notes}} in {{Computer Science}}, Springer Nature Switzerland (2022)

\bibitem{mask3d}
Schult, J., Engelmann, F., Hermans, A., Litany, O., Tang, S., Leibe, B.: {Mask3D: Mask Transformer for 3D Semantic Instance Segmentation}  (2023)

\bibitem{hugginggpt}
Shen, Y., Song, K., Tan, X., Li, D., Lu, W., Zhuang, Y.: Hugginggpt: Solving ai tasks with chatgpt and its friends in huggingface. arXiv preprint arXiv:2303.17580  (2023)

\bibitem{ProgPrompt}
Singh, I., Blukis, V., Mousavian, A., Goyal, A., Xu, D., Tremblay, J., Fox, D., Thomason, J., Garg, A.: {{ProgPrompt}}: {{Generating Situated Robot Task Plans}} using {{Large Language Models}}. In: 2023 {{IEEE International Conference}} on {{Robotics}} and {{Automation}} ({{ICRA}}). pp. 11523--11530 (2023)

\bibitem{vipergpt}
Sur\'is, D., Menon, S., Vondrick, C.: Vipergpt: Visual inference via python execution for reasoning. ICCV  (2023)

\bibitem{bbh}
Suzgun, M., Scales, N., Scharli, N., Gehrmann, S., Tay, Y., Chung, H.W., Chowdhery, A., Le, Q.V., hsin Chi, E.H., Zhou, D., Wei, J.: Challenging big-bench tasks and whether chain-of-thought can solve them. In: ACL (2022)

\bibitem{cot}
Wei, J., Wang, X., Schuurmans, D., Bosma, M., Xia, F., Chi, E., Le, Q.V., Zhou, D., et~al.: Chain-of-thought prompting elicits reasoning in large language models. NeurIPS  \textbf{35} (2022)

\bibitem{MP3D-EQA}
Wijmans, E., Datta, S., Maksymets, O., Das, A., Gkioxari, G., Lee, S., Essa, I., Parikh, D., Batra, D.: Embodied question answering in photorealistic environments with point cloud perception. In: CVPR (2019)

\bibitem{visualchatgpt}
Wu, C., Yin, S., Qi, W., Wang, X., Tang, Z., Duan, N.: Visual chatgpt: Talking, drawing and editing with visual foundation models (2023)

\bibitem{objectnet3d}
Xiang, Y., Kim, W., Chen, W., Ji, J., Choy, C., Su, H., Mottaghi, R., Guibas, L., Savarese, S.: Objectnet3d: A large scale database for 3d object recognition. In: ECCV (2016)

\bibitem{clevr3d}
Yan, X., Yuan, Z., Du, Y., Liao, Y., Guo, Y., Li, Z., Cui, S.: Comprehensive visual question answering on point clouds through compositional scene manipulation. arXiv preprint arXiv:2112.11691  (2021)

\bibitem{3d-bonet}
Yang, B., Wang, J., Clark, R., Hu, Q., Wang, S., Markham, A., Trigoni, N.: Learning object bounding boxes for 3d instance segmentation on point clouds. NeurIPS  \textbf{32} (2019)

\bibitem{llm-grounder}
Yang, J., Chen, X., Qian, S., Madaan, N., Iyengar, M., Fouhey, D.F., Chai, J.: Llm-grounder: Open-vocabulary 3d visual grounding with large language model as an agent (2023)

\bibitem{MM-REACT}
Yang, Z., Li, L., Wang, J., Lin, K., Azarnasab, E., Ahmed, F., Liu, Z., Liu, C., Zeng, M., Wang, L.: {{MM-REACT}}: {{Prompting ChatGPT}} for {{Multimodal Reasoning}} and {{Action}} (2023)

\bibitem{3dqa}
Ye, S., Chen, D., Han, S., Liao, J.: 3d question answering. IEEE Trans Vis Comput Graph  (2022)

\bibitem{zero-grounding}
Yuan, Z., Ren, J., Feng, C.M., Zhao, H., Cui, S., Li, Z.: Visual programming for zero-shot open-vocabulary 3d visual grounding (2023)

\bibitem{FE-3DGVQA}
Zhao, L., Cai, D., Zhang, J., Sheng, L., Xu, D., Zheng, R., Zhao, Y., Wang, L., Fan, X.: Towards explainable 3d grounded visual question answering: A new benchmark and strong baseline. IEEE Trans Circuits Syst Video Technol  (2022)

\bibitem{leasttomost}
Zhou, D., Sch{\"a}rli, N., Hou, L., Wei, J., Scales, N., Wang, X., Schuurmans, D., Cui, C., Bousquet, O., Le, Q.V., Chi, E.H.: Least-to-most prompting enables complex reasoning in large language models. In: ICLR (2023)

\bibitem{MiniGPT-4}
Zhu, D., Chen, J., Shen, X., Li, X., Elhoseiny, M.: {{MiniGPT-4}}: {{Enhancing Vision-Language Understanding}} with {{Advanced Large Language Models}} (2023)

\bibitem{3d-vista}
Zhu, Z., Ma, X., Chen, Y., Deng, Z., Huang, S., Li, Q.: 3d-vista: Pre-trained transformer for 3d vision and text alignment. ICCV  (2023)

\end{thebibliography}

\clearpage
\setcounter{section}{0}
\setcounter{table}{0}
\setcounter{figure}{0}

\renewcommand{\thetable}{S\arabic{table}}
\renewcommand{\arraystretch}{0.8}

\renewcommand{\thefigure}{S\arabic{figure}}
\renewcommand{\arraystretch}{0.8}

\section*{\centering Appendix}

The supplementary material is organized as follows.
In~\cref{sec:model_details}, We present details of our model including 3D visual perception module (\cref{subsec:visual_modules}) and prompt construction for the preparation stage and reasoning stage (\cref{subsec:prompts}).
Then we describe the details of evaluation protocals and the compared baselines in~\cref{sec:evaluation_metrics}.
Finally, we provide additional results and failure case analysis in~\cref{sec:cases}.

\section{Model Details}
\label{sec:model_details}

\subsection{3D Visual Perception Module}
\label{subsec:visual_modules}

\subsubsection{3D Object Segmentation.}
We use Mask3D~\cite{mask3d} trained on ScanNet200~\cite{ScanNet200} dataset to generate instance segmentations for the scene point clouds. 

\subsubsection{3D Object Classification.}
\label{sec:perc_seg}
We explore two settings for object category classification: using 607 categories (r1-r4 in Table 3) v.s. using 200 categories (r5 in Table 3).
For r5 in Table 3, we directly use the 200-category labels corresponding to instance segmentations generated by Mask3D~\cite{mask3d} trained on ScanNet200~\cite{ScanNet200}.
For r3 and r4 in Table 3, we adopt a hierarchical approach to classify objects into one of 607 categories in order to better satisfy SQA3D which covers 607 raw categories to answer the questions.
Specifically, we map the 607-category labels from groundtruth segmentations and the 200-category labels from Mask3D predicted segmentations to NYU40 classes using the label mapping table provided by ScanNet~\cite{scannet}.
We first classify an object point cloud into one of the broader high-level categories (\eg "table") which consist of NYU40 classes.
Then, we further classify the object into subcategories within the corresponding high-level category (\eg "coffee table", "dining table") via OpenShape~\cite{openshape}.
We extract the 3D features using the pre-trained 3D model OpenShape~\cite{openshape} and then fit a KNN model for each high-level category on ScanNet train set to classify the object-centric features into subcategory labels.
The KNN classification achieves an overall accuracy of 77.8\% when evaluated on SQA3D test set scenes using groundtruth segmentations.

\subsubsection{Attribute classification.}
\label{sec:perc_attr_cls}

We explore two approaches for attribution classification: using human-annotated captions from ScanRefer~\cite{scanrefer} (r2 in \cref{tab:ablation_context}) v.s. using the pre-trained OpenShape model (r1 in \cref{tab:ablation_context}).

When using captions, the LLM needs to interpret and extract relevant information indirectly from the captions. For example, to identify the color of the chair, the LLM should first call the $query\_caption$ API to get the caption of the chair, \eg "It is a black chair, near a brown table.". Then the LLM extracts "black" from the caption as the color of the chair.

When using OpenShape for attribute classification, we feed the colored point cloud and attribute candidates to OpenShape. The OpenShape model uses a 3D encoder to extract 3D features from the point cloud, and a text encoder to extract text features for each candidate attribute. We measure the cosine similarity between the 3D feature and the text feature of each attribute candidate, and select the one with the highest similarity as the final result.

\subsubsection{Spatial relation recognition.}
\label{sec:perc_spatial}

We define our spatial relations by adapting on the 4 types of relations, including horizontal (closest, farthest, \etc), vertical (above, on, \etc), allocentric (left, right, \etc), and between, in Nr3D from Referit3D~\cite{referit3d}.  We opt out of the ternary relation type, since the 'between' relation is infrequent in our task. Specifically, there are three types of spatial relations in our work:
\begin{itemize}
    \item horizontal relations: 'closest', 'farthest', 'within reach', 'around', \etc.
    \item vertical relations: 'on', 'above', 'below', \etc.
    \item allocentric relations: 'left', 'right', 'front', 'back', \etc.
\end{itemize}

\noindent\textbf{Horizontal relations.} Besides 'closest' and 'farthest' as in Nr3D, we add additional 'within reach' and 'around' given the characteristics of SQA3D.

 1) For 'closest' and 'farthest', we calculate the distance between the target object $O_t$ and all other objects using NearestNeighbors, and sort the objects according their distances into an ascended list $[O_1, O_2, \dots, O_{N-1}]$, where $N$ is the total number of objects in the scene. if $dist(O_1, O_t) + \epsilon < dist(O_2, O_t)$, then $O_1$ is considered 'closest' to $O_t$. Similarly, if $dist(O_{N-2}, O_t) + \epsilon < dist(O_{N-1}, O_t)$, then $O_{N-1}$ is considered the 'farthest' to $O_t$.

 2) For 'within reach' and 'around', given a target object $O_t$ and an anchor object $O_a$, if $dist(O_t, O_a) < wr\_dist $, then the target is within reach of the anchor. Similarly, if $dist(O_t, O_a) < ar\_dist $, then the target is around the anchor.
Here, $\epsilon$, $wr\_dist$ and $ar\_dist$ are pre-defined hyper-parameters.

\noindent\textbf{Vertical relations} include 'on', 'above', and 'below'. Given a target object $O_t$ and an anchor object $O_a$, we use their 2D IOU and distance in the z-axis to predict their vertical spatial relation. Take the relation 'on' as an example, the prediction is calculated as follows. Let $O_t^{2D}$ and $O_a^{2D}$ denote their 2D bounding box in the XY plane. If $IOU(O_t^{2D}, O_a^{2D}) < \text{min\_iou}$, then they have no vertical relations. If the following conditions are satisfied, then the target is 'on' the anchor, a.k.a. they have the 'on' vertical relation. 

\begin{align}
    \frac{\text{intersect}(O_t^{2D}, O_a^{2D})}{\text{area}(O_a^{2D})} & > \text{min\_on\_ratio} \\
    \left| z^{bottom}_t - z^{top}_a \right| & \leq \text{max\_on\_dist} \\
    \frac{\text{area}(O_t^{2D})}{\text{area}(O_a^{2D})} & < \text{max\_on\_ratio}
\end{align}

\noindent\textbf{Allocentric relations} include 'left', 'right', 'front' and 'back'. Take the 'left' relation as an example, we first define a polygon that represents the region to the left of the anchor object. Then we calculate the spatial overlap between the target object and the polygon. If the overlap exceeds a threshold, then the 
target object is on the 'left' of the anchor, otherwise they have no 'left' relation. All other allocentric relations can be derived in a similar fashion.

\subsection{Prompt Construction}
\label{subsec:prompts}
\subsubsection{Prompts for Preparation Stage.}
The prompt applied during the Preparation Stage contains four parts, including task definition, format specification, API documentation and in-context examples.

\noindent
\textbf{Task definition.} As shown in \cref{lst:task_definition}, the task definition prompt instructs the LLM to solve the situated reasoning task adhering to a predefined output format.

\noindent
\textbf{Format specification.} As shown in \cref{lst:format_specification}, the format specification prompt guides the LLM to provide a response with either "program" or "final answer".

\noindent
\textbf{API documentation.} As shown in \cref{lst:api_documentation},
the LLM is allowed to utilize the functions specified in the API documentation during the process of program generation.

\noindent
\textbf{In-context examples.} As shown in \cref{lst:example}, in-context examples are demonstrated to teach the LLM how to engage in an iterative Think-Program-reCtify loop.

\subsubsection{Prompts for Reasoning Stage.}
We define different prompts for various scenarios, including program execution failure, response parsing error and summarization need when the maximum number of attempts is reached.

\noindent
\textbf{Prompts for program execution failure.} As shown in \cref{lst:execution_failure}, if the program execution fails, the LLM is prompted with a debug
command to loop back to the Think phase and adjust the plan and program accordingly.

\noindent
\textbf{Prompts for parsing error.} As shown in \cref{lst:parsing_error}, if the LLM’s response does not adhere to the required format, it is prompted
with an instruction to loop back to the Think phase and make adjustments to the plan and program accordingly.

\noindent
\textbf{Prompts for summarization.} As shown in \cref{lst:summary}, if the maximum number of attempts is reached, the LLM is instructed to generate the final answer based on known observations.

\noindent
\textbf{Prompts for ensemble.} As shown in \cref{lst:ensemble}, we merge the top 5 answers and probabilities from an end-to-end model and the open-ended responses from our LLM-TPC, and feed them into another LLM to generate the final answer.

\section{Evaluation Protocals and Baselines}
\label{sec:evaluation_metrics}

\subsection{Evaluation Protocols}

\subsubsection{Soft match.}
The "\emph{soft match}" protocol takes into account the inherent ambiguity in natural language expressions and allows for partial credit when a model's predictions are close to the groundtruth but not identical.
To be specific, we first clean the predicted answer and groundtruth answer by removing punctuation, special characters, and converting Arabic numerals to their corresponding English words. 
Then we use the $soft\_match$ function (defined in \cref{lst:soft_em}) to compare the predicted answer with the groundtruth answer. It checks for different conditions to determine if they match, including exact equality, sub-string containment, word similarity, common word presence, and synonym detection (defined in \cref{tab:syn_expr}). If any of these conditions are met, the function returns $True$; otherwise, it returns $False$.

\subsubsection{Strict match.}

SQA3D~\cite{sqa3d} adopts "\emph{strict match}" as the evaluation metric, \ie the accuracy of answer classification given 706 candidate answers in the test set.
However, \emph{strict match} is not reasonable in cases where the open-ended generation from LLM does not exactly match the groundtruth answer, even though it conveys the same meaning (as shown in the first 3 rows of \cref{tab:strict_soft}). Additionally, the \emph{strict match} metric may not account for situations where multiple answers are correct but not annotated (as shown in the last 3 rows of \cref{tab:strict_soft}).
Therefore, we adopt the "\emph{soft match}" metric to evaluate LLM-generated answers in the main paper: if the predicted answer has an intersection with the groundtruth words, or if they comply with our predefined synonym rules, the prediction is deemed as correct.

\begin{figure}[h]
\begin{lstlisting}[language=Python,style=style-python]
def soft_match(answer, gt_answer):
    # remove punctuation and special characters, convert Arabic numerals to their corresponding English words
    answer = clean_answer(answer)
    gt_answer = clean_answer(gt_answer)
    if answer == gt_answer:
        return True
    elif answer in gt_answer:
        return True
    elif gt_answer in answer:
        return True
    elif ''.join(answer.split()) in ''.join(gt_answer.split()):
        return True
    elif ''.join(gt_answer.split()) in ''.join(answer.split()):
        return True
    elif len(set(answer.split()).intersection(gt_answer.split())) > 0:
        return True
    elif is_synonym(answer, gt_answer):
        return True
    return False
\end{lstlisting}
\mycaption{Soft match protocol}{The soft match protocol takes into account the inherent ambiguity in natural language expressions and allows for partial credit when a model's predictions are close to the groundtruth but not identical. It checks for different conditions to determine if they match, including exact equality, substring containment, word similarity, common word presence, and synonym detection. If any of these conditions are met, the function returns $True$; otherwise, it returns $False$.}
\label{lst:soft_em}
\vspace{-10pt}
\end{figure}

\begin{table}[t]
    \small
    \caption{Synonymous expressions in answers.}
    \label{tab:syn_expr}
    \vspace{-10pt}
    \begin{tabular}{>{\centering\arraybackslash}m{0.14\linewidth}|>{\arraybackslash}m{0.33\linewidth}||>{\centering\arraybackslash}m{0.14\linewidth}|>{\arraybackslash}m{0.33\linewidth}}
         \toprule
         answer & \multicolumn{1}{c||}{synonymous expressions} & answer & \multicolumn{1}{c}{synonymous expressions} \\
         \midrule
\rowcolor{gray!20} left & left, 7 o'clock, 8 o'clock, 9 o'clock, 10 o'clock, 11 o'clock &
right & right, 1 o'clock, 2 o'clock, 3 o'clock, 4 o'clock, 5 o'clock \\
front & front, forward, forwards, in front, infront, 10 o'clock, 11 o'clock, 12 o'clock, 1 o'clock, 2 o'clock &
behind & behind, back, backward, backwards, 4 o'clock, 5 o'clock, 6 o'clock, 7 o'clock, 8 o'clock \\
\rowcolor{gray!20} true & true, yes &
false & false, no \\
big & big, large &
circle & circle, circular, oval, round \\
\rowcolor{gray!20} rectangle & rectangle, rectangular &
box & box, boxes \\
cabinet & cabinet, cabinets &
chair & chair, chairs \\
\rowcolor{gray!20} clothes dryer & clothes dryer, clothes dryers &
clothing & clothing, clothes \\
cube & cube, cubes &
curtain & curtain, curtains \\
\rowcolor{gray!20} divider & divider, dividers &
dryer & dryer, dryers \\
kitchen cabinet & kitchen cabinet, kitchen cabinets &
mail box & mail box, mail boxes \\
\rowcolor{gray!20} mini fridge & minifridge, mini fridge &
monitor & monitor, monitors \\
picture & picture, pictures &
pillow & pillow, pillows \\
\rowcolor{gray!20} pipe & pipe, pipes &
plant & plant, plants \\
rack & rack, rack stand &
towel & towel, towels \\
\rowcolor{gray!20} trash bin & trash bin, trash bins, trash can, trashcan &
washing machine & washing machine, washing machines \\
whiteboard & whiteboard, white board &
window & window, windows \\
         \bottomrule
    \end{tabular}
\end{table}

\begin{table}[H]
    \centering
    \footnotesize
    \caption{Examples where LLM-TPC's prediction is evaluated as correct under "\emph{soft match}" while incorrect under "\emph{strict match}". Strict match is not reasonable when the predicted answer does not strictly match the groundtruth answer (GT) despite conveying the same meaning, or multiple answers are correct but not annotated.}
    \label{tab:strict_soft}
    \vspace{-8pt}
    \begin{tabular}{>{\arraybackslash}m{0.6\linewidth}|>{\centering\arraybackslash}m{0.12\linewidth}|>{\centering\arraybackslash}m{0.15\linewidth}}
    \toprule
        \centering Question & GT & Prediction\\
        \midrule
        \rowcolor{gray!20} What shape is the table behind me? & rectangle & rectangular \\
        What direction should I walk if I wanted to head towards the door to exit the room? & forward & front \\
        \rowcolor{gray!20} Is the toilet seat covered up or down to my left? & up & covered up \\
        If I turn my head, what could I see above the bed behind me? & TV & picture, TV \\
        \rowcolor{gray!20} What would make a splashing sound I pressed on a handle to my right? & toilet & sink, toilet \\
        What color is the bag to my right? & red & black, red \\
            \bottomrule
    \end{tabular}
\end{table}

\subsection{Implementation Details of the Baselines}

\noindent \textbf{ScanQA~\cite{sqa3d}.} In the setting w/o GT, ScanQA processes the entire 3D scan and extracts object-centric features using VoteNet~\cite{votenet}. We retrain the model as conducted in the original paper. In the setting w/ GT, ScanQA is provided with groundtruth object labels, RGB colors, location, and size features for retraining. Following ViL3DRel~\cite{vil3dref}, (i) we encode the groundtruth object labels using pretrained GloVe word vectors~\cite{glove};  (ii) we fit a Gaussian Mixture model on the RGB values of all points in the object. We set the mixture component to 3, linearly project the mean value in each component into a color embedding, and then use the weighted average as the final color embedding; (iii) we obtain the location features by linearly projecting the object center and size. Finally, the feature for each object consists of the concatenation of its object labels, RGB colors, location and size features.

\noindent \textbf{3D-VisTA~\cite{3d-vista}, LEO~\cite{leo} and LLM-TPC.} In the setting w/o GT, we apply segmentation masks predicted by Mask3D~\cite{mask3d} trained on ScanNet200~\cite{scannet} covering 200 categories. 
In the setting w/ GT, we apply groundtruth segmentation masks covering 607 categories.

\section{Additional Results}
\label{sec:cases}

\subsection{Quantitative Comparisons}

We present the answer accuracy under the "\emph{strict match}" protocol in \cref{tab:compare_strict}.
End-to-end trained models, whether tackling the 3D-SR task through close-vocabulary classification (ScanQA~\cite{sqa3d} and 3D-VisTA~\cite{3d-vista}) or autoregressive generation (3D-LLM~\cite{3dllm} and LEO~\cite{leo}), are specifically trained or finetuned on the SQA3D dataset. This enables them to learn  patterns of answer distributions in the dataset, resulting in better performance for strict matching evaluation.
In contrast, LLM-TPC addresses the 3D-SR task without any additional training. It relies on its pre-existing knowledge base and language generation capabilities. As a result, LLM-TPC lacks familiarity with patterns of the dataset and does not have good scores for strict matching. The brief answers generated by LLM-T, adhering to instruction constraints, better align with the concise nature of SQA3D answers, resulting in better performance under the \emph{strict match} metric compared to LLM-TPC.

\begin{table}[h]
\vspace{-10pt}
\centering
\footnotesize
\caption{Answer accuracy (\%) on the full test set of SQA3D. "GT" denotes the groundtruth information. "Soft-m" and "Strict-m" denote answer accuracy under "\emph{soft match}" and "\emph{strict match}", respectively.}
\label{tab:compare_strict}
\vspace{-8pt}
\begin{tabular}{ll|ccc|cc} \toprule
 & \multirow{2}{*}{Method} & \multicolumn{3}{c|}{w/o GT}  & \multicolumn{2}{c}{w/ GT}\\
\cmidrule(r){3-7} 
 & & Proposal & Soft-m & Strict-m  &  Soft-m & Strict-m \\ \midrule
\multirow{4}{*}{Supervised} & ScanQA~\cite{sqa3d} & VoteNet~\cite{votenet} & 46.38 & 44.33 & 47.74 & 45.64 \\
 & 3D-VisTA~\cite{3d-vista} & Mask3D~\cite{mask3d} & 50.44 & 47.43 & 50.72 & 47.60 \\
 & 3D-LLM~\cite{3dllm} & - & 50.21 & 48.11 & - & - \\
 & LEO~\cite{leo} & Mask3D~\cite{mask3d} & 52.83 & \textbf{49.64} & 53.25 & \textbf{49.36} \\ \midrule
\multirow{2}{*}{Few-shot} & LLM-T & - & - & - & 48.28 & 43.22 \\
 & LLM-TPC (Ours) & Mask3D~\cite{mask3d} & 45.84 & 27.11 & 56.92 & 36.49 \\ \midrule
Ensemble & LLM-TPC + 3D-VisTA & Mask3D~\cite{mask3d} & \textbf{55.81} & 40.78 & \textbf{58.20} & 42.46 \\ \bottomrule
\end{tabular}
\vspace{-10pt}
\end{table}

\subsection{Qualitative Examples}

\begin{figure}[h]
    \centering
    \includegraphics[width=\linewidth]{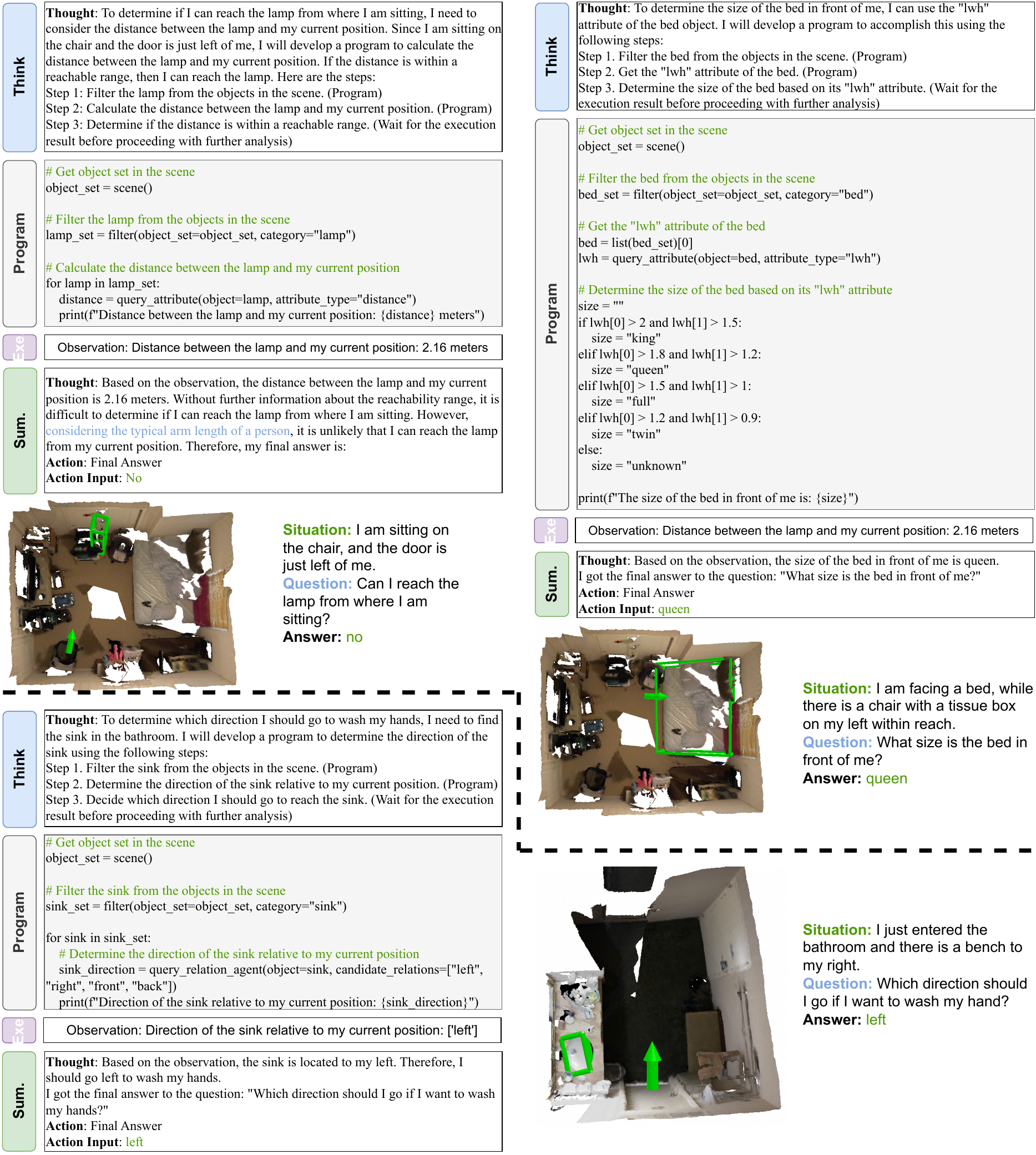}
        \mycaption{Additional qualitative results}{LLM-TPC can leverage common-sense reasoning to provide knowledge-intensive answers that go beyond the information derived solely from the 3D scene. For example, it can determine whether it can reach the lamp from its current position by considering the typical arm length of a person (as shown in the top-left figure). It can also estimate the size of the bed based on its length and width (as shown in the top-right figure). Furthermore, it can make decisions about which furniture can be used for washing hands (as shown in the bottom figure).}
        \label{fig:success1}
\end{figure}

\begin{figure}[h]
    \centering
    \includegraphics[width=\linewidth]{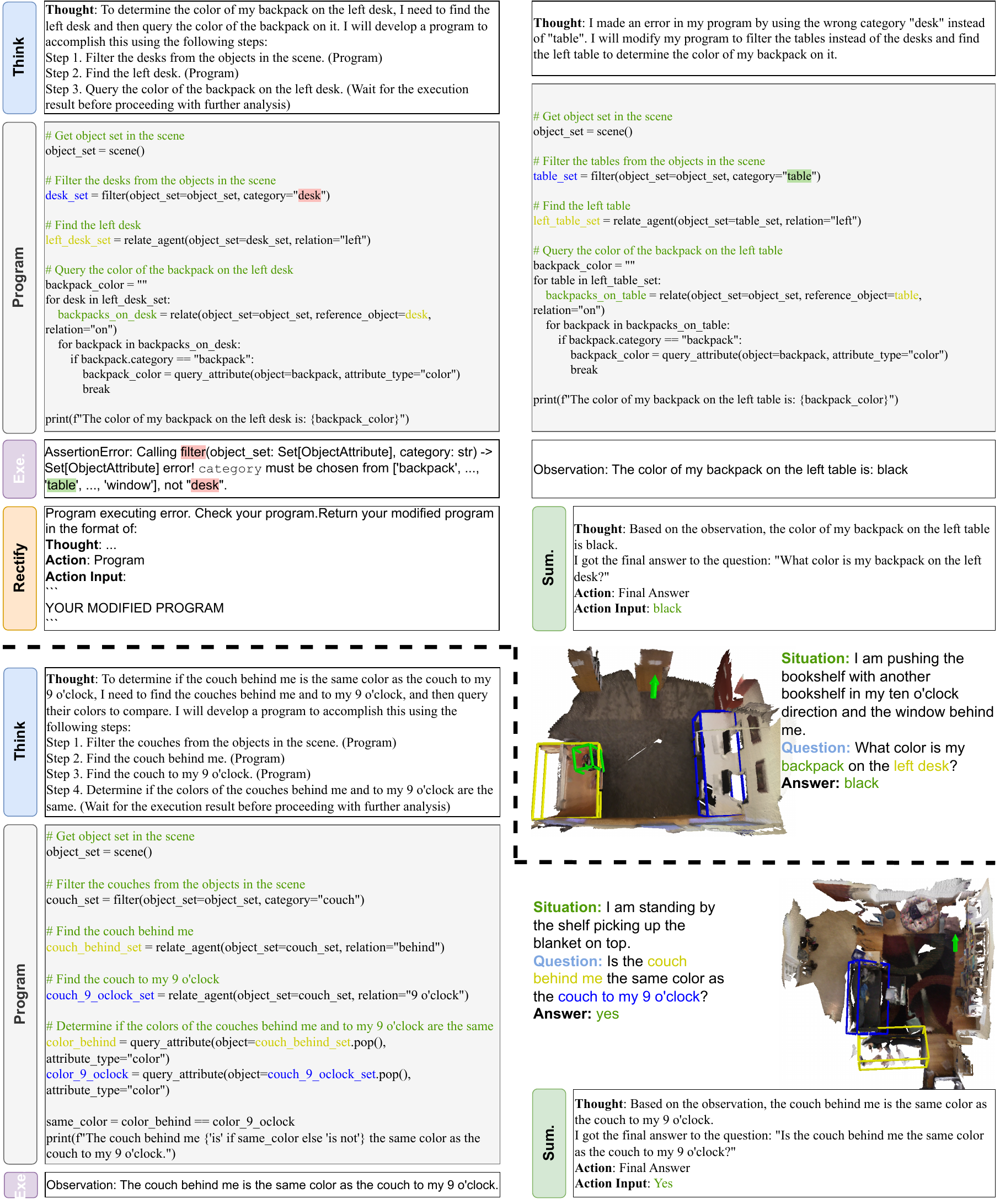}
        \mycaption{Additional qualitative results}{When the execution fails, LLM-TPC enters the Rectify module, where the LLM is prompted with a debug command to loop back to the Think phase and adjusts the plan and program according to the received error message (as shown in the top figure). Additionally, we can detect the capabilities of APIs and foundational visual models by observing the intermediate output of program execution (as shown in the bottom figure).}
        \label{fig:success2}
\end{figure}

\begin{figure}
    \centering
    \includegraphics[width=\linewidth]{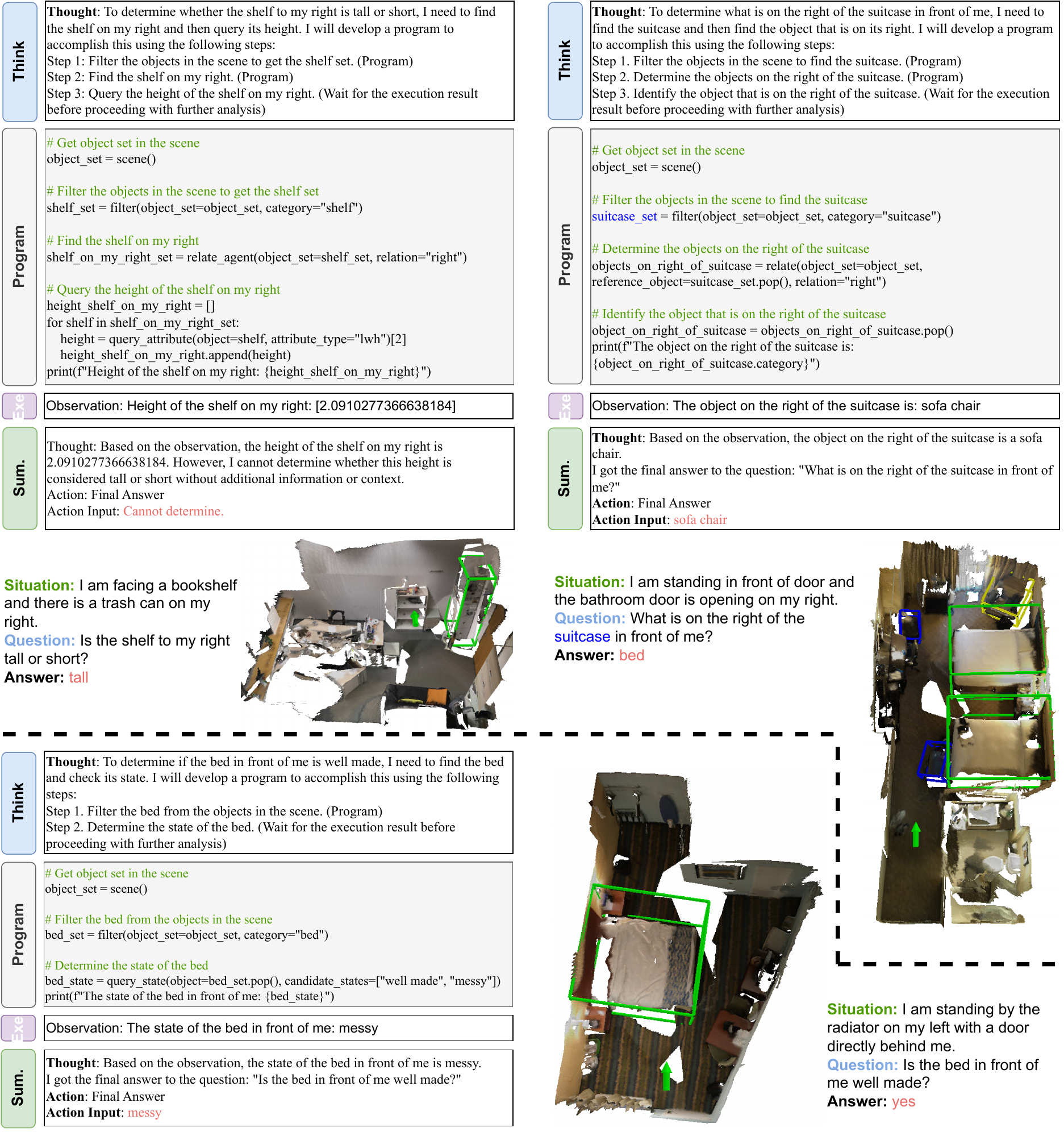}
        \mycaption{Failure cases}{In the top-left figure, without additional information or context (\eg the height of a human), it can be challenging for LLM-TPC to determine whether the shelf is tall or short based solely on the 3D scene. In the top-right figure, there are two suitcases in front of the agent and both the beds and the sofa chair are on the right side of the suitcases. In the bottom figure, the efficacy of LLM-TPC is highly dependent on the visual perception models used for attribute identification.}
        \label{fig:fail}
\end{figure}

We provide additional qualitative results in \cref{fig:success1,fig:success2} and failure cases in \cref{fig:fail}.

LLM-TPC can leverage common-sense knowledge to reason the knowledge-intensive answers that go beyond the information derived solely from the 3D scene. For example, it can determine whether it can reach the lamp from its current position by considering the typical arm length of a person (as shown in the top-left figure in \cref{fig:success1}). It can also estimate the size of the bed based on its length and width (as shown in the top-right figure in \cref{fig:success1}). Furthermore, it can make decisions about which furniture can be used for washing hands (as shown in the bottom figure in \cref{fig:success1}).

When the program execution fails, LLM-TPC enters the Rectify module, where the LLM is prompted with a debug command to loop back to the Think phase and adjusts the plan and program according to the received error message (as shown in the top figure in \cref{fig:success2}). Additionally, we can detect the capabilities of APIs and foundational visual models by observing the intermediate output of program execution (as shown in the bottom figure in \cref{fig:success2}).
LLM-TPC showcases interpretability with human-readable plans and programs while also demonstrating robustness through error rectification, plan adjustments based on error messages, and detecting API capabilities using intermediate outputs.

Failure cases in LLM-TPC can be attributed to several factors. Firstly, a lack of contextual information makes it challenging for the system to determine the final answer accurately (as shown in the top-left figure in \cref{fig:fail}). Secondly, the presence of multiple objects belonging to the same category can create ambiguity, leading to failure in understanding spatial relationships between objects (as shown in the top-right figure in \cref{fig:fail}). Lastly, inaccurate attribute identification by visual perception models can result in decreased system performance (as shown in the bottom figure in \cref{fig:fail}).

\newpage

\begin{figure}[h]
\begin{lstlisting}[language=Python,style=style-prompt]
You are a smart embodied agent. Use your coding and common sense reasoning skills to solve a question answering task with interleaving Thought, Action, Observation steps. Given your situation and question, use the following format to solve the task:
Thought: Answer the question by reasoning about scene and your situation. If you need further information about the objects in the scene (e.g. spatial relationship), generate a plan step by step and implement it in a program.
Action: The action to take, should be one of [Final Answer, Program].
Action Input:
(1) For Final Answer, return the answer to the question with NO MORE THAN 3 words.
(2) For Program, generate a Python program according to your thought to help you understand the scene.
\end{lstlisting}
\mycaption{Task definition}{The LLM is instructed to solve the situated reasoning task adhering to a predefined output format.}
\label{lst:task_definition}
\end{figure}

\begin{figure}[h]
\begin{lstlisting}[language=Python,style=style-prompt]
Valid format for Final Answer
-----------------------------
Thought: Your reasoning process for the final answer. 
Action: Final Answer
Action Input: Your final answer with NO MORE THAN 3 words. (Use your common sense reasoning skills to infer missing information and give a specific final answer.)

Valid format for Program
------------------------
Thought: Your plan for further information about the objects in the scene.
Action: Program
Action Input:
```Python
YOUR PROGRAM (Use ```print(variable_value_you_want_to_know)``` to display the value of a variable.)
```
\end{lstlisting}
\mycaption{Format specification}{The LLM is expected to provide either a "program" or a "final answer" in its response.}
\label{lst:format_specification}
\end{figure}

\begin{figure}[h]
\ContinuedFloat*
\begin{lstlisting}[language=Python,style=style-prompt]
When generating a program, each object is represented as an instance of ObjectAttribute and you can use the following functions:
```Python
class ObjectAttribute:
    category: str # category of the object
    xyz: List[float] # center coordinates of the object

scene() -> Set[ObjectAttribute]:
    """
    Returns a set of objects in the scene.
    """

filter(object_set: Set[ObjectAttribute], category: str) -> Set[ObjectAttribute]:
    """
    Returns a set of objects whose category is `category`.

    Examples
    --------
    >>> # Get object set in the scene
    >>> object_set = scene()
    >>> # Filter all the tables
    >>> table_set = filter(object_set=object_set, category="table")
    """

relate_agent(object_set: Set[ObjectAttribute], relation: str) -> Set:
    """
    Returns a set of objects that are related to the agent(you) by the relation.

    Examples
    --------
    >>> # Find the table on my left
    >>> table_left_set = relate_agent(object_set=table_set, relation="left")
    """    

relate(object_set: Set[ObjectAttribute], reference_object: ObjectAttribute, relation: str) -> Set:
    """
    Returns a set of objects that are related to the reference_object by the relation.

    Examples
    --------
    >>> # Find objects on top of the table on my left
    >>> objects_on_table = set()
    >>> for table in table_left_set:
    >>>     objects_on_table.update(relate(object_set=object_set, reference_object=table, relation="on"))

    >>> # Determine what objects are on top of the table
    >>> objects_on_table_category = []
    >>> for obj in objects_on_table:
    >>>     objects_on_table_category.append(obj.category)
    >>> print(f"Objects on top of the table on my left: {objects_on_table_category}")
    Objects on top of the table on my left: ['book', 'tray']
    """
\end{lstlisting}
\end{figure}

\begin{figure}[h]
\ContinuedFloat
\begin{lstlisting}[language=Python,style=style-prompt,firstnumber=52]
query_relation_agent(object: ObjectAttribute, candidate_relations: Optional[List[str]]=["left", "right", "front", "back", "o'clock"]) -> List:
    """
    Returns a list of allcentric relations between the object and the agent(you).
    If `candidate_relations` is provided, only relations in the `candidate_relations` list will be returned.

    Examples
    --------
    >>> # Decide which direction I should go to reach the table
    >>> direction = query_relation_agent(object=table)
    >>> print(f"Direction of the table relative to my current position: {direction}")
    >>> print(f"I should go {' '.join(direction)} to reach the table.")
    Direction of the table relative to my current position: ['left', 'back']
    I should go left back to reach the table.

    >>> # Decide whether the table is in front of me or behind
    >>> direction = query_relation_agent(object=table, candidate_relations=["front", "behind"])
    >>> print(f"Direction of the table relative to my current position: {' '.join(direction)}")
    Direction of the table relative to my current position: behind
    """

query_relation(object: ObjectAttribute, reference_object: ObjectAttribute, candidate_relations: Optional[List[str]]=["left", "right", "front", "back"]) -> List:
    """
    Returns a list of allcentric relations between the object and the reference_object.
    If `candidate_relations` is provided, only relations in the `candidate_relations` list will be returned.

    Examples
    --------
    >>> relation = query_relation(object=chair, reference_object=table)
    >>> print(f"The chair is in the direction of {' '.join(relation)} to the table")
    The chair is in the direction of left front to the table

    >>> relation = query_relation(object=chair, reference_object=table, candidate_relations=["left", "right"])
    >>> print(f"The chair is on the {' '.join(relation)} of the table")
    The chair is on the left of the table
    """
\end{lstlisting}
\end{figure}

\begin{figure}[h]
\ContinuedFloat
\begin{lstlisting}[language=Python,style=style-prompt,firstnumber=87]
query_attribute(object: ObjectAttribute, attribute_type: str, candidate_attribute_values: Optional[List[str]]) -> Union[List[float], float, str]:
    """
    Returns the attribute of the object.
    `attribute_type` must be chosen from the following list: ["lwh", "distance", "color", "shape", "material"].
    If `candidate_attribute_values` is provided, only values in the `candidate_attribute_values` list will be returned.

    Examples
    --------
    >>> lwh = query_attribute(object=object, attribute_type="lwh") # unit: meter. length, width, height of the object bounding box (unit: meter). Can be used to compute the length(lwh[0]), width(lwh[1]), height(lwh[2]), area(lwh[0]*lwh[1]) and volume(lwh[0]*lwh[1]*lwh[2]) of the object. Helpful for deciding the size of the object.
    >>> print(lwh)
    [0.68883693 0.29695976 0.17185348]
    
    >>> distance = query_attribute(object=object, attribute_type="distance") # unit: meter. Helpful for getting the distance of an object from the agent(you). Can be used to compare which object is closer or farther to the agent(you).
    >>> print(distance)
    2.3456789

    >>> # Determine whether the color of the object is brown, black or red
    >>> color = query_attribute(object=object, attribute_type="color", candidate_attribute_values=["brown", "black", "red"])
    >>> print(color)
    brown

    >>> # Determine whether the shape of the object is round, square or rectangular
    >>> shape = query_attribute(object=object, attribute_type="shape", candidate_attribute_values=["round", "square", "rectangular"])
    >>> print(shape)
    rectangular

    >>> # Determine whether the material of the object is wood or metal
    >>> material = query_attribute(object=object, attribute_type="material", candidate_attribute_values=["wood", "metal"])
    >>> print(material)
    wood
    """

query_state(object: ObjectAttribute, candidate_states: List[str]) -> str:
    """
    Returns the state of the object.

    Examples
    --------
    >>> state = query_state(object=object, candidate_states=["neat", "messy"])
    >>> print(state)
    neat
    """
```
\end{lstlisting}
\mycaption{API documentation}{During the Program phase, the LLM has the flexibility to utilize the functions defined in the API documentation.}
\label{lst:api_documentation}
\end{figure}

\begin{figure}[h]
\begin{lstlisting}[language=Python,style=style-prompt]
//user:// 
I am in a room. Looking around me, I see some objects: 2 chair, 1 coffee table, 1 couch, 2 lamp, 1 pillow, 2 table, 2 window.
My situation: I am facing a chair, while having a couch on my right and a coffee table behind me.
Question: What is behind me directly?

[[assistant:]] 
Thought: Based on the information provided, I need to find the object behind me and close to me. I will develop a program to determine what is directly behind me using the following steps:
Step 1. Identify objects behind me. (Program)
Step 2. Sort the objects behind me by distance. (Program)
Step 3. Determine what object is directly behind me. (Wait for the execution result before proceeding with further analysis)
Action: Program
Action Input:
```Python
# Get object set in the scene
object_set = scene()

# Identify objects behind me
object_behind_set = relate_agent(object_set=object_set, relation="behind")

# Sort the objects behind me by distance
object_behind_by_distance = list(object_behind_set)
object_behind_by_distance.sort(key=lambda x: query_attribute(x, "distance"))

# Determine what object is directly behind me
category_behind_by_distance = [obj.category for obj in object_behind_by_distance][:3]
print(f"Objects directly behind me: {category_behind_by_distance}")
```

//user:// 
Observation: Objects directly behind me: ['coffee table', 'couch', 'pillow']

[[assistant:]]
Thought: Based on the observation, there are multiple objects behind me in proximity: coffee table, couch, pillow. Considering my situation, I am facing a chair, while having the couch on my right and the coffee table behind me. Therefore, the coffee table is located directly behind me.
I got the final answer to the question: "What is behind me directly?"
Action: Final Answer
Action Input: coffee table
\end{lstlisting}
\mycaption{Exemplar in-context example}{The LLM engages in an iterative Think-Program-reCtify loop, generating plans, developing programs using API calls to implement the plan, and obtaining answers based on program execution results. If the program executes successfully, it summarizes the final answer.}
\label{lst:example}
\end{figure}

\begin{figure}[h]
\begin{lstlisting}[language=Python,style=style-prompt]
Program executing error. Check your program. |ERROR INFORMATION| Return your modified program in the format of:
Thought: ...
Action: Program
Action Input:
```Python
YOUR MODIFIED PROGRAM
```
\end{lstlisting}
\mycaption{Prompts for program execution failure in the Rectify phase}{If the program execution fails, the LLM is prompted with a debug command to loop back to the Think phase and adjust the plan and program accordingly. "\textcolor{red}{ERROR INFORMATION}" will be replaced with the actual error information during the execution of the program.}
\label{lst:execution_failure}
\end{figure}

\begin{figure}[h]
\begin{lstlisting}[language=Python,style=style-prompt]
Response parsing error. Check your response and return your response in the format of:
Thought: ...
Action: The action to take, should be one of [Final Answer, Program].
Action Input: ...

Valid format for Final Answer
-----------------------------
Thought: Your reasoning process for the final answer. 
Action: Final Answer
Action Input: Your final answer with NO MORE THAN 3 words. (Use your common sense reasoning skills to infer missing information and give a specific final answer.)

Valid format for Program
------------------------
Thought: Your plan for further information about the objects in the scene.
Action: Program
Action Input:
```Python
YOUR PROGRAM (Use ```print(variable_value_you_want_to_know)``` to display the value of a variable.)
```
\end{lstlisting}
\mycaption{Prompts for parsing error in the Rectify phase}{If the LLM's response does not adhere to the required format, it is prompted with an instruction to loop back to the Think phase and make adjustments to the plan and program accordingly.}
\label{lst:parsing_error}
\end{figure}

\begin{figure}[h]
\begin{lstlisting}[language=Python,style=style-prompt]
You have reached the maximum number of chats. Anyway, still try your best to give a final answer in the format of:
Thought: Reasoning about the objects in the scene, your situation and question, you can still give a final answer. Use your common sense reasoning skills to infer missing information and give a specific final answer.
Action: Final Answer
Action Input: your final answer with NO MORE THAN 3 words.
\end{lstlisting}
\mycaption{Prompts for summarization}{If the maximum number of attempts is reached, the LLM is instructed to generate the final answer based on known observations.}
\label{lst:summary}
\end{figure}

\begin{figure}[h]
\begin{lstlisting}[language=Python,style=style-prompt]
Consider a 3D environment where various objects are present. You are provided with a question related to this environment, and your task is to reason about the situation and provide reasonable answers.

You have two sources of information to help you answer this question:

1. A large language model has produced an open-ended answer based on its strong reasoning capabilities:
   - LLM's answer: [Open-ended answer here]

2. An end-to-end model has predicted the following top 5 answers with their associated probabilities:
   - Answer 1: [Probability for Answer 1]
   - Answer 2: [Probability for Answer 2]
   - ...
   - Answer 5: [Probability for Answer 5]

Using your advanced reasoning skills, evaluate both the close-vocabulary answers from the end-to-end model and the open-ended answer from the large language model. Consider the context of the question, the likelihood of each answer, and the plausibility based on your situation in the 3D environment. Provide your answers following the format of the examples, with no additional words. Always try your best to give the reasonable answers.

Example 1:
Question: I am sitting on the chair. Which direction should I go if I want to leave the room?
LLM's answers with their associated probabilities:
 - left front: 1.0
End-to-end model's top 5 answers with their associated probabilities:
 - left: 0.89
 - forward: 0.10
 - front: 0.02
 - door: 0.01
 - right: 0.01
Reasonable answers: left front forward

Example 2:
Question: I am standing in a room. Is the shape of table in front of me round, square or rectangular?
LLM's answers with their associated probabilities:
 - square: 1.0
End-to-end model's top 5 answers with their associated probabilities:
 - rectangular: 0.99
 - closed: 0.01
 - l-shaped: 0.01
 - round: 0.01
 - square: 0.01
Reasonable answers: rectangular

Example 3:
Question: I am facing a bed. What is on the table that is on my right?
LLM's answers with their associated probabilities:
 - book chair couch: 1.0
End-to-end model's top 5 answers with their associated probabilities:
 - monitor: 0.89
 - monitors: 0.05
 - printer: 0.01
 - plant: 0.01
 - TV: 0.01
Reasonable answers: book monitor

Question: {question}
LLM's answers with their associated probabilities:
 - {ans}: 1.0
End-to-end model's top 5 answer with their associated probabilities:
 - {ans1}: {prob1}
 - {ans2}: {prob2}
 - {ans3}: {prob3}
 - {ans4}: {prob4}
 - {ans5}: {prob5}
Reasonable answers: 
\end{lstlisting}
\mycaption{Prompts for ensemble}{We merge the top 5 answers and probabilities from an end-to-end model and the open-ended responses from our LLM-TPC, and feed them into another LLM to generate the final answer.}
\label{lst:ensemble}
\end{figure}

\end{document}